\documentclass[pdflatex,default,iicol,sn-mathphys]{sn-jnl}

\jyear{2023}%

\theoremstyle{thmstyleone}%
\theoremstyle{thmstyletwo}%

\theoremstyle{thmstylethree}%

\newcommand{\blankfootnote}[1]{%
  \refstepcounter{footnote}%
  \let\tempthefootnote\thefootnote
  \let\thefootnote\relax\footnotetext{#1}%
  \let\thefootnote\tempthefootnote
}

\usepackage{doi}
\usepackage{academicons}
\usepackage{hyperref}
\definecolor{orcidlogocol}{HTML}{A6CE39}
\begin{document}

\title[ ]{Intelligent Escape of Robotic Systems: A Survey of  Methodologies, Applications, and Challenges}


\author[]{\fnm{Junfei} \sur{Li} }\email{jli64@uoguelph.ca}

\author*[]{\fnm{Simon} \spfx{X.} \sur{Yang}\textsuperscript{*}}\email{syang@uoguelph.ca}

\affil[ ]{\orgdiv{School  of Engineering}, \orgname{University  of Guelph}, \orgaddress{\street{50 Stone Road East}, \city{Guelph}, \postcode{N1G2W1}, \state{ON}, \country{Canada}}}

\abstract{ Intelligent escape is an interdisciplinary field that employs artificial intelligence (AI) techniques to enable robots with the capacity to intelligently react to potential dangers in dynamic, intricate, and unpredictable scenarios. As the emphasis on safety becomes increasingly paramount and advancements in robotic technologies continue to advance, a wide range of intelligent escape methodologies has been developed in recent years. This paper presents a comprehensive survey of state-of-the-art research work on intelligent escape of robotic systems. Four main methods of intelligent escape are reviewed, including planning-based methodologies, partitioning-based methodologies, learning-based methodologies, and bio-inspired methodologies. The strengths and limitations of existing methods are summarized. In addition, potential applications of intelligent escape are discussed in various domains, such as search and rescue, evacuation, military security, and healthcare. In an effort to develop new approaches to intelligent escape, this survey identifies current research challenges and provides insights into future research trends in intelligent escape.}




\maketitle

\section{Introduction}\label{sec1}
\begin{figure*}[h]
\centering
\includegraphics[width=0.75\textwidth]{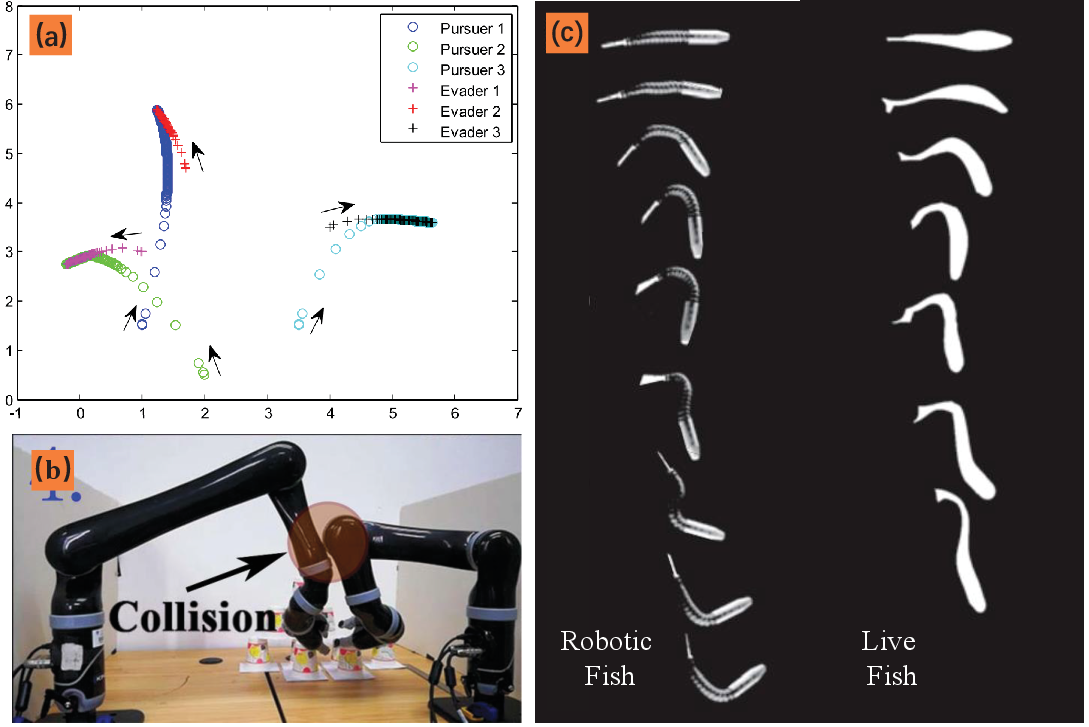}
\caption{Intelligent escape combines knowledge and methodologies from multiple disciplines or subject areas. (a) robot evasion strategies in pursuit-evasion games \cite{lopez2019solutions}; (b) collision avoidance of robot
manipulators \cite{zhang2020mutual}; (c) animal-inspired escape behaviors \cite{currier2020bio}.}
\label{fig_examples}
\end{figure*}

\blankfootnote{This paper has been accepted into Journal of Intelligent \& Robotic Systems. \doi{https://doi.org/10.1007/s10846-023-01996-y}}
Intelligent escape is a multidisciplinary area in the field of robotics. 
Intelligent escape combines knowledge and methodologies from multiple disciplines or subject areas that allow robots to move quickly and effectively away from potentially dangerous or harmful situations.
As shown in Fig. \ref{fig_examples}, the intelligent escape combines methodologies from the design of evasion strategies for robots in pursuit-evasion games \cite{casini2022two,lopez2019solutions,xu2022multiplayer}, collision avoidance in robot systems \cite{yasin2020unmanned,zhang2020mutual,ding2020collision}, or the emulation of animal escape behavior found in nature \cite{li2016dynamics,currier2020bio}.
The scope of the problem extends beyond escape behavior itself and encompasses research in a number of other fields. Over the years, there has been a significant amount of research to develop an effective and efficient escape for robots. The research has primarily focused on creating sophisticated methodologies, algorithms, and techniques that can improve the safety, reliability, and agility of robots in uncertain or dangerous environments. 

In early studies, most robot escapes relied on pre-programmed escape routines \cite{boyer2021towards}. The pre-programmed routines were designed to allow robots to navigate away from specific hazards or emergency situations, such as avoiding obstacles or escaping from a fire. The pre-programmed routines are able to provide a simple solution in an unchanged environment. However, the pre-programmed routines are not able to provide a real-time response in unexpected or complex environments.  Researchers also explored sensor-based methods to implement robot escape \cite{birajdar2020development}. The sensor-based methods rely on sensor information to detect potential hazards, such as changes in temperature or the presence of toxic gases, and then make decisions. However, the early sensor-based methods have limitations to handle complex environments due to the scarcity and limited precision of available sensors.

In recent years, as the capabilities of robots and their sensors have improved, researchers have begun to explore more advanced techniques to implement robot escape \cite{muhammad2019designing,haddeler2020evaluation,jin2021bioinspired}. Intelligent escape is an emerging and thriving field that uses artificial intelligence (AI) techniques to enable robots with the intelligent ability to respond to potential dangers in more dynamic, complex, and uncertain environments. For instance, in the pursuit-evasion game, performance functions are used to represent the interest of the pursuers in reaching the position of the evaders and the interest of the evaders in avoiding capture. As shown in Fig. \ref{fig_examples}(a), these individual performance indices are then used to define the Nash equilibrium and other approaches \cite{lopez2019solutions}. In addition, as shown in Fig. \ref{fig_examples}(b),  a novel recurrent neural network was proposed to avoid the mutual collision of dual robot manipulators while performing collaboration tasks \cite{zhang2020mutual}. Furthermore, recent advances in nature-inspired robotics technology have also led to the infusion of biological intelligence into robotic systems to develop new forms of robot escape. As shown in Fig. \ref{fig_examples}(c), mimicking fish escape could provide insights into the design of novel robot escape \cite{currier2020bio}. Overall, intelligent escape represents a promising direction in the field of robotics, as the development of intelligent escape allows robots to operate more effectively and efficiently in uncertain and dynamic environments. With continued advancements in AI techniques, it is expected that intelligent escape will play an increasingly important role in the design of robots for a wide range of applications.

This paper aims to comprehensively survey the state-of-the-art research on intelligent escape behaviors for robotic systems. The focus of this survey is on four critical methods of intelligent escape, including planning-based methodologies, partitioning-based methodologies, learning-based methodologies, and bio-inspired methodologies. 
This survey also aims to critically evaluate the strengths and limitations of current intelligent escape research. In addition, this survey provides potential applications and insights into future developments of robot escape.

The organization of the paper is as follows: Section \ref{method} elucidates the state-of-the-art research on intelligent escape.  Section \ref{sec:application} provides the potential application of robotic escape.  Section \ref{sec:challenges} discusses the challenges and future research. In Section \ref{sec:conclusion}, the survey is briefly summarized.

\section{Current Studies of Intelligent Escape}
\label{method}
The objective of intelligent escape is to provide robotic systems with safety-conscious escape solutions. This field has witnessed substantial development in numerous research domains, such as path planning, machine learning, pursuit-evasion games, and collision avoidance.  In recent years, there is a trend to use biologically inspired approaches to face more complex situations.  In this section, a synopsis of the four principal methodologies is provided, along with their respective merits and constraints.

\subsection{Planning-based Methodologies}
\label{Planning}

The planning-based methodology is one kind of promising methodology, which involves the use of advanced planning techniques, algorithms, and approaches to control and guide the escape of the robot. These methodologies have gained significant attention in recent years and have become an active area of research in the field of robotics.  Table \ref{method_pathplanning} summarizes the current studies of planning-based methodologies.

\begin{table*}[h]   
\centering
\caption{Current Studies of Planning-based Methodologies}
\label{method_pathplanning}   
\resizebox{\linewidth}{!}{
\begin{tabular}{cccccc}    
\toprule
  Reference& Main approach & Environment& Obstacle avoidance& Coordination& Sensing ability \\    
\midrule   
Gilles and Vladimirsky(2020) \cite{gilles2020evasive} & Dynamic programming &Land (2D)& Static & No &Limited\\  
Gong \textit{et al.}(2022) \cite{gong2022online} & Dynamic programming & Land (2D)& No & No &Perfect\\   
Zhou and Xu(2022) \cite{zhou2022decentralized} & Mean field games theory& Land (2D)& No& No & Perfect\\  
 Scott and Leonard(2018) \cite{scott2018optimal}  & State-feedback control & Land (2D)& No & Yes &Perfect\\  
  Scott \textit{et al.}(2018)\cite{lewis2018minimum}  & Pontryagin's minimum principle & Land (2D)& No &No&Perfect\\ 
Makkapati and Tsiotras(2019) \cite{makkapati2019optimal}  & Multi-objective optimization & Land (2D) &No &No&Perfect\\ 
  Hu \textit{et al.}(2019)  \cite{hu2019multiobjective}  & Multi-objective optimization & Water Surface (2D) &Static and dynamic &Yes&Limited\\ 
Wang \textit{et al.}(2021) \cite{wang2021path}  & APF and MPC &Land (2D)& Static & No&Perfect\\ 
Saravanakumar \textit{et al.}(2021) \cite{saravanakumar2021sampling}  & Modified RRT  & Air(3D) & Static & No&Perfect\\ 
Dong \textit{et al.}(2012)  \cite{dong2012strategies} & APF & Land (2D) & Static &No& Perfect\\ 
Chen \textit{et al.}(2020) \cite{chen2020dynamic} & A* search and improved APF & Air (3D) & Static and dynamic & No &Perfect\\ 
Ajeil \textit{et al.}(2021)  \cite{ajeil2021novel} &Bat swarm optimization & Land (2D)  & Dynamic &No&Limited\\
 Zaccone and Martelli(2020)   \cite{zaccone2020collision} & RRT & Water Surface (2D)  & Static and dynamic & Yes&Perfect\\ 
 Tang \textit{et al.}(2021)   \cite{tang2021gwo} & APF & Land (2D) & Static & Yes& Limited\\\ 
 Goodwin  \textit{et al.}(2015)  \cite{goodwin2015escape} & ACO &  Land (2D) & Static & No &Perfect\\ 
  Zhu  \textit{et al.}(2020)  \cite{zhu2020receding} & MPC &  Land (2D) & Static and dynamic & No &Perfect\\ 
Cognetti \textit{et al.}(2017) \cite{cognetti2017real} & LoS &  Land (2D) & No & No &Perfect\\
\bottomrule   
\end{tabular}  
}
\end{table*}

The dynamic programming method has been widely applied in robotics, including for the prevention of robot escapes. The dynamic programming method involves breaking a complex problem down into smaller, more manageable sub-problems and then solving them recursively. By applying dynamic programming to the problem of robot escape, it is possible to generate an optimal path for the robot to follow that minimizes the risk of escape. 
In the event that the optimal strategies, denoted as $\vec{u}_B^*(\vec{x})$ and $\vec{u}_R^*(\vec{x})$, are unique or selected from a set of optimal strategies, the game's dynamics, represented by $f_i\left(\vec{x}, \vec{u}_B, \vec{u}_R\right)$
can be evaluated by substituting these strategies. The resulting integral yields the optimal paths leading to the game's termination from a given set of initial conditions.
The value of the game satisfies the following equation \cite{isaacs1999differential}

\begin{equation}
\sum_i \frac{\partial V}{\partial x_i} f_i\left(x, \hat{u}_B, \hat{u}_R\right)+G\left(x, \hat{u}_B, \hat{u}_R\right)=0
\label{diff1}
\end{equation}
where $V$ and $G$ are the value and payoff of the game, respectively. Variables $\vec{u}_B$ and $\vec{u}_R$ are control inputs of the evader and pursuer, respectively. Variables $\hat{u}_B$ and $\hat{u}_R$ are estimates of the evader and pursuer, respectively. The optimal strategies are expressed as functions of both  $\hat{x}$ and $\frac{\partial V}{\partial x_i}$, and they are obtained by evaluating 
$\hat{u}_B$ and $\hat{u}_R$ once $ V_(\vec{x})$ is known.  By differentiating with respect to each, the following results are obtained \cite{isaacs1999differential}
\begin{equation}
\frac{d}{d t} \frac{\partial V}{\partial x_k}=-\left(\sum_i \frac{\partial V}{\partial x_i} \frac{\partial f_i}{\partial x_k}+\frac{\partial G}{\partial x_k}\right)
\label{diff2}
\end{equation}
and applying $\vec{u}_B^*(\vec{x})$ and $\vec{u}_R^*(\vec{x})$ to the game dynamics gives
\begin{equation}
\dot{\vec{x}}=f\left(\vec{x}, \vec{u}_B^*(\vec{x}), \vec{u}_R^*(\vec{x})\right)
\label{diff3}
\end{equation}
where $\vec{x}$ is the state vector, which contains the locations and possibly the headings of the players. The complex dimensionality poses a significant challenge in the numerical solution of games. This is particularly evident in the case of fully nonlinear, partial differential equations such as Hamilton-Jacobi equations, where the complexity of the problem increases exponentially with the number of dimensions.
Gilles and Vladimirsky \cite{gilles2020evasive} incorporated multi-objective dynamic programming to design evasive paths under surveillance uncertainty.
The proposed approach is able to guide robots in static-obstacle environments.  
Gong \textit{et al.} \cite{gong2022online} proposed an online adaptive dynamic programming method  to enable each agent to obtain the policy for reaching the Nash equilibrium.
The proposed approach is able to generate the optimal path. However, the environment is assumed a collision-free environment.

The  optimization approaches are also used in the field of intelligent escape. Zhou and Xu \cite{zhou2022decentralized} developed a decentralized approach to designing optimal evasion strategies in a large-scale pursuit-evasion game using the Mean Field Games theory approach. 
The proposed approach is capable of dealing with large-scale problems that are not considered in many studies. However, the proposed approach only considers a collision-free and full-knowledge environment, which may not meet the real-world requirement. 
Makkapati and Tsiotras \cite{makkapati2019optimal}  proposed a novel algorithm that combines game theory and multi-objective optimization to determine the optimal strategy for evaders.
The proposed approach provided an optimal solution to evasion. However, the generated evasive path is not flexible and adaptive to complex and dynamic environments.
Scott and Leonard \cite{scott2018optimal} proposed a strategy for evaders that was not initially targeted to avoid capture and considered the limited sensing condition through a local strategy of risk reduction. 
However, the control design of robots may not deal with the large-number  cooperation problem. 
Scott \textit{et al.} \cite{lewis2018minimum} discussed the minimum-time evasion trajectories for evaders with motion constraints. 
However, the proposed minimum-time trajectory did not consider the influence of the obstacle.
Hu \textit{et al.} \cite{hu2019multiobjective} presented a multi-objective optimization approach for planning evasive paths of autonomous surface vehicles. However, there are some possible risks that the convention is subject to various interpretations causing uncertainty. 


Path planning is another promising technique for preventing robot escapes.
In path planning-based escape methodologies, the starting position corresponds to the current location of the robot, while the target position is identified as a possible escape point. The robot is required to generate the shortest path that avoids obstacles to reach the escape point. Therefore, following the collision-free path generated, the robot can remain within its designated operating space and avoid collisions and maintain human safety by identifying humans as obstacles. Saravanakumar \textit{et al.}\cite{saravanakumar2021sampling} presented a sampling-based evasion path planning algorithm for unmanned aerial vehicles (UAVs) to avoid collisions with moving obstacles or other UAVs.
 However, the proposed approach has an inability to handle dynamic obstacles and has only limited coordination ability.
 Ajeil \textit{et al.} \cite{ajeil2021novel} proposed a modified bat swarm optimization for the evasion of dynamic obstacles. However, this paper did not discuss the potential benefits of using heuristic-based methods, such as their ability to yield satisfactory convergence.  Zaccone and Martelli \cite{zaccone2020collision} proposed a rapid exploration random tree (RRT)-based evasive maneuver that is able to avoid collision with the obstacle.
 However, the proposed approach did not provide enough solutions to the local minimum problem when using RRT-based approaches.
 Goodwin  \textit{et al.} \cite{goodwin2015escape} proposed a near-optimal escape plan, based on Ant Colony Optimisation (ACO), for every affected person in considering the dynamic spread of fires, moving impairments caused by hazards, and faulty unreliable data. 
 However, the computing complexity might become expensive when considering large-number robots.
 Zhu  \textit{et al.} \cite{zhu2020receding} used the model predictive control (MPC) method to obtain evasion strategies that estimate control variables in real-time to solve evasion problems with obstacles present.
 However, the proposed approach has some limitations when considering the influence of model errors. MPC control is based on the mathematical model of the system, and if there are errors in the model, the control effect may be affected.

 The artificial potential field (APF) constitutes a significant approach to designing escape mechanisms for robotic systems. The APF technique models the environment as a potential field, where attractive and repulsive forces guide the navigation of the robot. Attractive forces are generated by the target destination, while repulsive forces originate from obstacles, steering the robot away from potential collisions. Taking advantage of the interaction of these forces, the robot can efficiently navigate toward the goal while avoiding obstacles, thereby facilitating effective escape strategies. 
 Wang \textit{et al.} \cite{wang2021path} proposed an APF method to generate a collision-free evasive path, and MPC to design the path planner. The proposed approach is capable of providing individually safe trajectories for different drivers with maneuverability on large curvature roads. 
However, the proposed approach might need a high computational requirements, which require complex mathematical calculations to solve the optimal control strategy, which would require high computation power and time.
 Dong \textit{et al.} \cite{dong2012strategies} incorporated the APF method to design a collision-free evasion model. 
 However, the proposed approach did not consider the deadlock situation, where the evader robots might be trapped in a narrow and complex obstacle environment. 
 Chen \textit{et al.} \cite{chen2020dynamic} presented the UAV dynamic evasive trajectory planning using the sparse A * search and the improved APF method to avoid the sudden static threat and the moving obstacle. 
 However, the sparse A* search and the improved APF method might not provide a real-time response to dynamic changes in the environment.
 Tang \textit{et al.} \cite{tang2021gwo} proposed a prey escape strategy based on APF repulsion. When the distance between the robot and the target decreases, the repulsion force increases, and the pursuit robot can avoid local optima situations.
 However, the proposed approach did not consider a narrow and complex obstacle environment.

The planning-based methodologies provide an efficient solution to intelligent escape. However, many studies might not perform the ideal performance when considering model constraints or real-time responses in real-world applications. For instance, the humanoid robot needs to deal with the gait generation problem, which means that the pure feedback scheme cannot be used. Cognetti \textit{et al.} \cite{cognetti2017real} proposed a real-time planning procedure to generate the evasion trajectory of the humanoid robot. The control law for the evader is designed as \cite{cognetti2017real}
\begin{equation}
\begin{aligned}
v_e  =-\bar{v} 
\end{aligned}
\end{equation}
\begin{equation}
\begin{aligned}
\omega_e  =k\left(\theta_{\mathrm{eva}}-\theta_e\right)
\end{aligned}
\end{equation}
where $\bar{v}$ is the constant driving velocity of the pursuit behavior. Variables $v_e $ and $\omega_e$ are the input of the driving and steering speed of the evader, respectively. Parameter $k$ is a positive constant. Variable $\theta_e$ is the current direction of the evader. Variable $\theta_{\mathrm{eva}}=\angle \boldsymbol{n}_{\mathrm{eva}}-\pi$, where $\boldsymbol{n}_{\mathrm{eva}}$ is the unit vector representing the chosen direction for evasion. Variable $\angle \boldsymbol{n}_{\mathrm{eva}}$ is the phase angle of $\boldsymbol{n}_{\mathrm{eva}}$. In their study, the choice of $\boldsymbol{n}_{\mathrm{eva}}$ encodes the chosen evasion strategy. In this paper, there are two strategies, indicated by \textit{move back} and \textit{move aside} strategies. The strategy \textit{move back} is that the evader determines the line-of-sight to the pursuer, represented by $\boldsymbol{n}_{\mathrm{eva}}$. The strategy \textit{move aside} is that the evader moves backward so as to align with a direction that is orthogonal to the line-of-sight to the pursuer. This corresponds to the setting $ \boldsymbol{n}_{\mathrm{eva}}=\boldsymbol{n}^{\perp}_{\mathrm{aim}}$, where $\boldsymbol{n}^{\perp}_{\mathrm{aim}}$ is the normal unit vector to the line-of-sight to the evader in the half-plane behind the robot. The proposed approach entailed embracing a re-planning methodology, which is predicated on the current understanding of the other robot and promptly modifies this plan to accommodate new perceptions.

 As mentioned in the literature reviews, the current state of planning-based methodologies is analyzed as follows:
 \begin{enumerate}
\item  In terms of obstacle avoidance, some research only concentrates on static obstacles (e.g., \cite{gilles2020evasive,wang2021path,saravanakumar2021sampling,dong2012strategies,tang2021gwo,goodwin2015escape}), while only a few of the research considers both static and dynamic obstacles (e.g., \cite{hu2019multiobjective,chen2020dynamic,zaccone2020collision,zhu2020receding}). This trend indicates that contemporary research is progressively tackling the more challenging problem of dynamic environments.  
\item Although some studies have made progress in multi-robot coordination (e.g., \cite{scott2018optimal,hu2019multiobjective,zaccone2020collision,tang2021gwo}), many have not addressed the coordination issue (e.g., \cite{gilles2020evasive,gong2022online,zhou2022decentralized,lewis2018minimum,makkapati2019optimal,wang2021path,saravanakumar2021sampling,dong2012strategies,chen2020dynamic,ajeil2021novel,goodwin2015escape,zhu2020receding,cognetti2017real}). This suggests that cooperative obstacle avoidance by multiple robots could be a vital direction for future research.
\item Regarding the sensing capabilities, many studies assume robots work with perfect perception (e.g., \cite{gong2022online,zhou2022decentralized,scott2018optimal,lewis2018minimum,makkapati2019optimal,wang2021path,saravanakumar2021sampling,dong2012strategies,chen2020dynamic,zaccone2020collision,goodwin2015escape,zhu2020receding,cognetti2017real}), while only a few studies consider limited sensing abilities (e.g., \cite{gilles2020evasive,hu2019multiobjective,ajeil2021novel,tang2021gwo}). This observation implies that researchers are focused on developing effective avoidance strategies under limited sensing conditions.
\end{enumerate}
The planning-based methodologies are able to improve escape performance in various environments and scenarios. Future trends suggest a potential increase in future investigations into multi-robot cooperative avoidance strategies and obstacle avoidance under limited sensing capabilities. 

\subsection{Partitioning-based Methodologies}
\label{Partitioning}

The partitioning-based methodologies divide the environment into regions with different levels of safety and importance. By dividing the environment in this way, these methods enable robots to make more informed decisions about their behavior and movements and can help them escape. Table \ref{method_Partitioning} summarizes the current studies of partitioning-based methodologies.

\begin{table*}[h]   
\centering
\caption{Current Studies of Partitioning-based Methodologies}
\label{method_Partitioning}   
\resizebox{\linewidth}{!}{
\begin{tabular}{cccccc}    
\toprule
  Reference& Main approach & Environment& Obstacle avoidance& Coordination&Sensing ability\\    
\midrule  
Liang \textit{et al.}(2023) \cite{liang2023collaborative} & Apollonius Circle & Air (2D)& Static& Yes&Perfect\\   
 Liang \textit{et al.}(2019) \cite{liang2019multi}  & Apollonius Circle & Land (2D)& No & Yes &Limited\\  
Ramana and Kothari(2017) \cite{ramana2017pursuit} &  Apollonius Circle & Land (2D)& No &No &Perfect\\  
Sun \textit{et al.}(2022) \cite{sun2022cooperative} &  Apollonius Circle& Land (2D)& Static &No &Perfect\\  
Kumar and Ojha(2019) \cite{kumar2019experimental}  & Apollonius Circle &Land (2D)& No & No&Perfect\\  
 Ruiz(2023) \cite{ruiz2023time} & Dominance region & Land (2D)& No & No&Limited\\ 
Oyler \textit{et al.}(2016) \cite{oyler2016pursuit}  & Dominance region &Land (2D)& Static &No &Perfect\\ 
Yan \textit{et al.}(2018) \cite{yan2018reach}   & Dominance region &Land (2D) &No &No&Perfect\\ 

\bottomrule   
\end{tabular}  
}
\end{table*}

The Apollonius Circle is a geometric construction that has important applications in robotics. This method involves the use of circles to represent the position and motion of a robot and can be used to generate optimal paths for the robot to follow. 
\begin{figure}[h]
\centering
\includegraphics[width=0.4\textwidth]{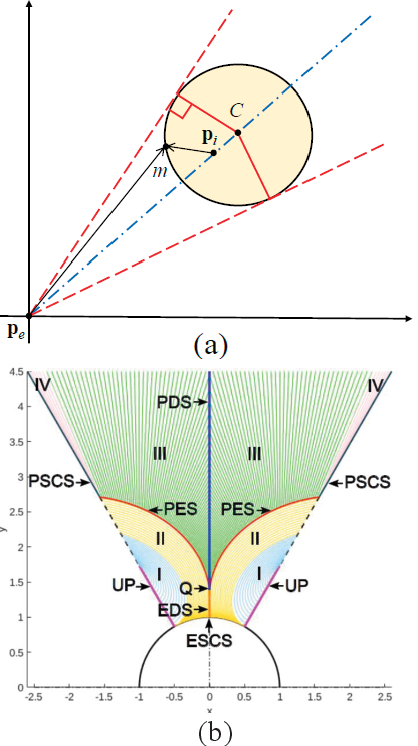}
\caption{Examples of the partitioning-based methodologies. (a) The illustration of the Apollonius Circle; (b) The illustration of the dominance regions \cite{ruiz2023time}. EDS: evader’s dispersal surface. ESCS: evader’s state constraint surface. PES: pursuer’s equivocal surface. PDS: pursuer’s dispersal surface. PSCS: pursuer’s state constraint surface. UP: usable part.} 
\label{AC}
\end{figure}
Suppose that both the pursuer and the evader begin to
move with their maximum speeds along fixed directions at
a time instant. If they meet at a point $m$ in a finite time, which is denoted by $m = (x_m, y_m)$, the point $m$ must satisfy the following equation \cite{fang2020cooperative}
\begin{equation}
\left(x_m-\frac{x_{i}}{1-\lambda^2}\right)^2+\left(y_m-\frac{y_{i}}{1-\lambda^2}\right)^2=\lambda^2 \frac{x_{i}^2+y_{i}^2}{\left(1-\lambda^2\right)^2}
\end{equation}
where $\lambda$ is the speed ratio of the pursuer and the evader. It is easy to conclude that $m = (x_m, y_m)$ is on a circle $\mathbf{C}=\left(\left[\left(x_{i}\right) /\left(1-\lambda^2\right)\right],\left[\left(y_{i}\right) /\left(1-\lambda^2\right)\right]\right)$ with radius $R=\sqrt{x_{i}^2+y_{i}^2\left[\left(\lambda\right) /\left(1-\lambda^2\right)\right]}$. This circle is known as the Apollonius Circle \cite{isaacs1999differential}, as shown in Fig. \ref{AC}(a). 
The speed ratio plays an important role in the evasion task \cite{fang2020cooperative}. If the speed ratio $\lambda_i$ is big, which means the evader can easily be captured by the pursuer. Moreover, if $\lambda<1$, which means the speed of the evader is greater than the pursuer, the evader can easily escape from the pursuer. Thus, in traditional considerations, there is an assumption that $\lambda>1$.
Since the speed ratio $\lambda >1$, thus the pursuer can capture all evaders in a finite time. Therefore, the purpose of evasion is to increase the total survival time of the evaders. Recently, many researchers have focused on the cooperative pursuit of the faster-moving evader \cite{fang2020cooperative}. Although the speed of the purser is much slower than the evader, the team of pursuers is able to capture the evader based on intelligent cooperation. This trend also indicates the potential research of intelligent escape against cooperative pursuit.

The Apollonius Circle method has been applied in a range of different robotic applications.   Liang \textit{et al.} \cite{liang2023collaborative} proposed a collaborative pursuit-evasion game based on the Apollonius Circle in an environment with obstacles. The proposed method uses a leader-follower mode and dynamic window approach to avoid obstacles and form formations. 
However, the proposed approach involves multiple calculation steps, each step may introduce errors, which may result in inaccurate results.
Liang \textit{et al.} \cite{liang2019multi} proposed an evasion solution using Apollonius circles to determine capture areas. They provided the possible escape angle for the robot. 
However, the proposed approach might not be applicable in some cases when the Apollonius Circle method may not find a solution. Ramana and Kothari \cite{ramana2017pursuit} proposed an escape strategy using the concept of the Apollonius Circle for a high-speed evader to escape a perfectly enclosed formation from multiple pursuers in an open domain.
Sun \textit{et al.} \cite{sun2022cooperative} proposed an intelligent evasion strategy in response to differing encirclement conditions, as well as the configuration of pursuer distribution predicated upon the Apollonian Circle.

\begin{figure*}[h]
\centering
\includegraphics[width=1\textwidth]{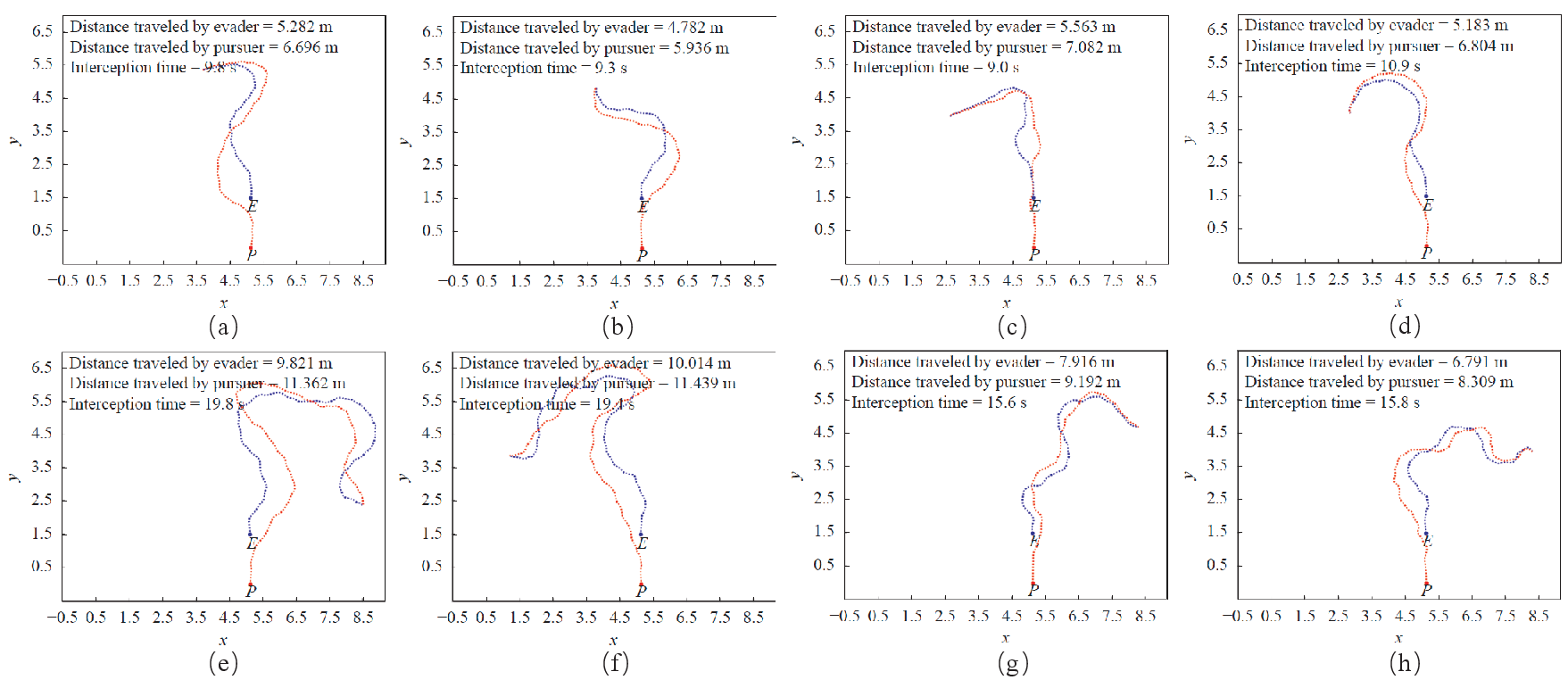}
\caption{The comparison of the classic evasion scheme and Apollonius circle-based evasion scheme \cite{kumar2019experimental}. (a) classical evasion against AIPNG;  (b) classical evasion against AAG; (c) classical evasion against modified AIPNG; (d) classical evasion against anticipated trajectory-based PNG; (e) Apollonius circle-based evasion against AIPNG;  (f) Apollonius circle-based evasion against AAG; (g) Apollonius circle-based evasion against modified AIPNG; (h) Apollonius circle-based evasion against anticipated trajectory-based PNG.}
\label{AC_compare}
\end{figure*}

The Apollonius Circle-based evasion is able to improve the evasion performance in many scenarios.
Kumar and Ojha \cite{kumar2019experimental} evaluated the performance of Apollonius Circle-based evasion and other strategies using mobile robots on wheels. As shown in Fig. \ref{AC_compare}, the classic evasion and Apollonius Circle-based evasion were compared against four different pursuit strategies, including augmented ideal proportional navigation guidance (AIPNG), angular acceleration guidance (AAG), modified AIPNG, and anticipated
trajectory-based PNG. 
Apollonius Circle-based evasion strategies exhibit a greater capacity to extend the duration of evasion in comparison with conventional evasion techniques.

 An alternative solution method that has been used makes use of dominance regions, where a point is said to be dominated by a player if they are able to reach that point before their opponent, regardless of the opponent’s actions.  Ruiz \cite{ruiz2023time} proposed a partition of the reduced space into four regions, each corresponding to particular time-optimal escape strategies. As shown in Fig. \ref{AC}(b), Region I represents the primary solution of the game, involving the maximum speed rotation of the pursuer (blue curves). Region II contains the ESCS and related trajectories (yellow curves), defining the PES's start point Q. Region III pertains to the PES (red curve) and trajectories (green curves), where the pursuer initially translates away from the evader at maximum speed before a combined motion. Region IV relates to the PSCS and its trajectories (pink curves), where the pursuer similarly starts with a backward translation at maximum speed, leading to a combined motion in realistic space. Dominance regions provide the complete solution to capture games with full information. 
 Oyler \textit{et al.} \cite{oyler2016pursuit} proposed dominance regions to generate the evasion path considering the presence of obstacles. 
 However, the proposed approach only considers the static obstacle. The dynamic obstacle might be a difficult problem due to the complex computation of the proposed approach.
 Yan \textit{et al.} \cite{yan2018reach} designed the attacker dominance region, which is able to guide the attacker to escape to a specified edge of the domain boundary. 
 However, the proposed approach did not consider the influence of the obstacle. In addition, similar to the Apollonius Circle method, the dominance regions introduce a significant computational load, escalating both the complexity and duration of the calculation.

As mentioned in the literature reviews, the current state of partitioning-based  methodologies is analyzed as follows:
 \begin{enumerate}
\item  Several studies do not consider obstacle avoidance (e.g., \cite{liang2019multi,ramana2017pursuit,kumar2019experimental,ruiz2023time,yan2018reach}), while only a few studies investigate static obstacle avoidance (e.g., \cite{liang2023collaborative,sun2022cooperative,oyler2016pursuit}). This might indicate that partitioning-based methodologies are more commonly employed in scenarios with minimal obstacles or that obstacle avoidance in partitioning-based methodologies requires further development.
\item Most of the research does not address the cooperation issue (e.g., \cite{ramana2017pursuit,sun2022cooperative,kumar2019experimental,ruiz2023time,oyler2016pursuit,yan2018reach}). This suggests that the future focus of partitioning-based methodologies could be the coordination of multi-robot systems.
\item Several studies assumed that the robot has complete information about the environment (e.g., \cite{liang2023collaborative,ramana2017pursuit,sun2022cooperative,kumar2019experimental,oyler2016pursuit,yan2018reach}). However, the assumption of perfect sensing capabilities in many studies may not be ideal for real-world applications.
\end{enumerate}
Partitioning-based methodologies currently focus on the Apollonius Circle and dominance region approaches, mainly in terrestrial 2D environments. Obstacle avoidance and coordination are not the main concerns of these methods but have been investigated in some specific studies. Furthermore, partitioning-based methodologies cater to both perfect and limited sensing capabilities, demonstrating adaptability to different systems and applications.

\subsection{Learning-based Methodologies}
\label{Learning}
\begin{table*}[h]   
\centering
\caption{Current Studies of Learning-based Methodologies}
\label{method_Learning}   
\resizebox{\linewidth}{!}{
\begin{tabular}{cccccc}    
\toprule
  Reference& Main approach & Environment& Obstacle avoidance& Coordination&Sensing ability\\    
\midrule  
Di \textit{et al.}(2019) \cite{di2019optimizing}  & Reinforcement learning & Land (2D) & No & No&Limited\\   
 Wang \textit{et al.}(2019) \cite{wang2019fuzzy} & Fuzzy and Reinforcement learning & Land (2D) & No & No & Perfect\\  
 Weitzenfeld(2008) \cite{weitzenfeld2008prey}   & Deep Reinforcement Learning & Land (2D) & No & No&Perfect\\ 
 Qi \textit{et al.}(2020) \cite{qi2020deep} &  Deep Reinforcement Learning & Land (2D) & Static &No &Perfect\\   
Zhang \textit{et al.}(2022)  \cite{zhang2022game}  & Deep
Reinforcement Learning & Air (3D)& Static & Yes&Limited\\  
Xu  \textit{et al.}(2022)  \cite{xu2022autonomous}  & Deep
Reinforcement Learning & Air (2D)& Static & No&Perfect\\ 
 Xu \textit{et al.}(2022) \cite{xu2022pursuit}  & Deep Reinforcement Learning & Land (2D) & No & No &Perfect\\ 
  Ji \textit{et al.}(2021)  \cite{ji2021obstacle} & Deep Reinforcement Learning & Land (2D) & Static and Dynamic & Yes&Perfect\\
 Gao \textit{et al.}(2019) \cite{gao2019fast}  & Meta-reinforcement learning &Land (2D) & Static &No &Limited\\ 
Shree  \textit{et al.}(2022)  \cite{shree2022learning}  & Bayesian fusion and  machine learning & Land (2D)  &No &Yes& Limited\\

\bottomrule   
\end{tabular}  
}
\end{table*}

The learning-based methodologies have shown great potential to prevent robot escapes. By leveraging machine learning techniques, this approach can enable robots to learn from their surroundings and make decisions that minimize the risk of escape. This approach has gained significant attention in recent years and is becoming an active area of research in the field of robotics. Table \ref{method_Learning} summarizes current studies on learning-based methodologies.

\begin{figure}[h]
\centering
\includegraphics[width=0.45\textwidth]{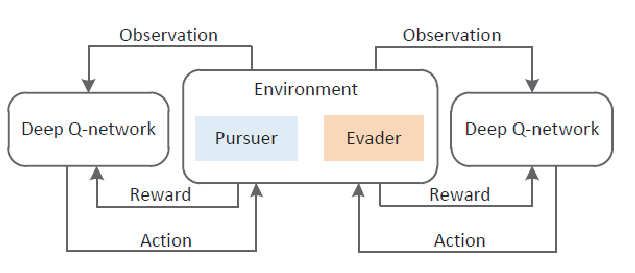}
\caption{Framework of the reinforcement learning for both pursuer and evader robots\cite{qi2020deep}.}
\label{self_play}
\end{figure}

\begin{figure*}[h]
\centering
\includegraphics[width=0.9\textwidth]{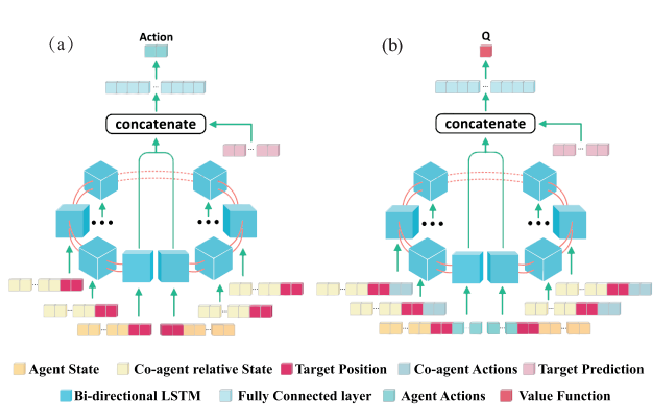}
\caption{Policy and critic network structures of CBC-TP Net. Quadcopter
agent corresponds to the initial and the final bidirectional LSTM cell, and
coagents correspond to other bidirectional LSTM cells \cite{zhang2022game}. (a) CBC-TP Net policy networks; (b) CBC-TP Net Q networks.}
\label{CBC-TP Net}
\end{figure*}

Reinforcement learning enables robots to learn optimal behaviors through trial-and-error interactions with their environment. Reinforcement learning advances robot adaptability, decision making, and autonomy, revolutionizing applications in manufacturing, service industries, and beyond. Fig. \ref{self_play} shows a classic self-play process based on reinforcement learning for both pursuer and evader robots. During the competitive process, the robots tend to engage with their opponents at a suitable level. Both robots can mutually improve each other to improve the diversity of trajectories, thereby enhancing the robustness of the performance. the learning process is able to avoid complex differential equations for escape scenarios. However, it is important to note that the modeling of the environment is an essential issue that can directly influence learning performance. In addition, the parameter selection in reinforcement learning algorithms has an important impact on model performance and effect, but these parameters often need to be manually adjusted, which requires the experience and knowledge of domain experts.
Di \textit{et al.} \cite{di2019optimizing} presented a memory-based reinforcement learning algorithm to optimize evasive strategies considering the limited field of view and noisy observation. 
However, the proposed approach did not explain how to deal with a large amount of training data. The testing may be required to collect data, which can be costly.
Wang \textit{et al.} \cite{wang2019fuzzy} proposed a novel fuzzy continuous reinforcement learning algorithm to increase the survival time of the evader. 
 However, the proposed approach might be difficult to handle in continuous state spaces. This may result in overly complex models that are difficult to train and optimize.
Weitzenfeld \cite{weitzenfeld2008prey} presented a novel neural-schema architecture for predator-avoidance tasks of multiple robots. The proposed architecture is based on deep reinforcement learning and swarm intelligence, allowing robots to adapt to their environment and cooperate to achieve their objectives. 
However, the proposed approach might result in poor performance when dealing with noise and uncertainty during the learning process.
Qi \textit{et al.} \cite{qi2020deep} applied a self-play mechanism to train evaders considering the impact of obstacles. Each robot was required to perform its evasion task while avoiding collisions with nearby obstacles. 
However, the proposed approach only considers the simple and static obstacle. The proposed approach might be difficult to handle in complex and dynamic environments.
Zhang \textit{et al.}  \cite{zhang2022game} proposed the pursuit-evasion framework to swarm aerial vehicles in the obstacle environment. They constructed multiagent coronal bidirectionally coordinated with target prediction network (CBC-TP Net) with a vectorized extension of multiagent deep deterministic policy gradient (MADDPG) formulation to ensure the effectiveness of the damaged swarm system, as shown in Fig. \ref{CBC-TP Net}.  Xu  \textit{et al.}  \cite{xu2022autonomous} proposed a novel deep reinforcement learning-based approach for autonomous obstacle avoidance and learning efficient navigation policies in complex environments.  
Xu \textit{et al.} \cite{xu2022pursuit} proposed the value-based deep Q network model and the deep deterministic policy gradient model for the dog sheep game, attempting to endow the sheep with the ability to escape successfully.  
However, the proposed approach might be difficult to handle in real-world applications because it only considers a collision-free environment.
Ji  \textit{et al.}  \cite{ji2021obstacle} developed a deep reinforcement learning-based algorithm that enables multiple agents to cooperate while maintaining their formation and avoiding obstacles in dynamic environments. 
  Both Xu  \textit{et al.}  \cite{xu2022autonomous} and Ji  \textit{et al.} \cite{ji2021obstacle}  have incorporated obstacle avoidance into their considerations, but the methodologies they propose might necessitate significant time and computational resources for model training.
Gao \textit{et al.} \cite{gao2019fast} used meta-reinforcement learning for fast adaptation to human-robot interactions in the escape room scenario. The proposed approach increased the perceived trustworthiness of the robot and influenced the gain of human trust in emergencies. Shree  \textit{et al.}  \cite{shree2022learning} proposed a multi-modal danger estimation pipeline for collaborative human-robot escape scenarios, which learning from the dataset of the movies and TV shows. 
However, the proposed approach might require a large amount of training data. The proposed approach might require a lot of experimentation and testing may be required to collect data, which can be costly.

As mentioned in the literature reviews, the current state of learning-based  methodologies is analyzed as follows:
 \begin{enumerate}
\item  Many studies do not consider dynamic obstacle avoidance (e.g., \cite{di2019optimizing,wang2019fuzzy,weitzenfeld2008prey,qi2020deep,zhang2022game,xu2022pursuit,xu2022autonomous,gao2019fast,shree2022learning}). This may imply that  future research could further investigate their potential in dynamic obstacle avoidance.
\item  Many studies do not consider the coordination issue (e.g., \cite{di2019optimizing,wang2019fuzzy,weitzenfeld2008prey,qi2020deep,xu2022pursuit,xu2022autonomous,gao2019fast}). This observation suggests that coordination is a vital element for future research.
\item Many studies assumed the robot has perfect sensing (e.g., \cite{wang2019fuzzy,weitzenfeld2008prey,qi2020deep,zhang2022game,xu2022pursuit}). However, perfect sensing may lead to reduced applicability in realistic environments. To improve the relevance of learning-based methodologies, researchers should emphasize the address and incorporation of limited sensing capabilities into future research.
\end{enumerate}
In summary, learning-based methodologies in robotics, particularly reinforcement learning, are gaining traction in the field. These approaches are predominantly applied to 2D land-based environments, with limited exploration in aerial and more complex settings. Although obstacle avoidance has not been extensively studied in learning-based methodologies, coordination has been investigated in several cases. Flexibility to work with limited and perfect sensing capabilities demonstrates the adaptability of learning-based methodologies to a variety of robotic systems and applications.

\subsection{Bio-Inspired Methodologies}

\label{Bio-Inspired}
Bio-inspired escape is a growing area of research that focuses on the development of robots that can respond to emergency situations in ways that are inspired by the behavior of living organisms \cite{Li2021bio}. The goal of bio-inspired escape is to create robots that can adapt to changing environments, navigate complex spaces, and respond to threats in ways that are similar to the way that animals respond to danger. To achieve this goal, researchers are studying the behavior of animals in emergency situations, such as how they navigate through their environment, how they detect and avoid danger, and how they coordinate their behavior with other members of their group. This information is then used to inform the design of robots that can perform similar behaviors. Table \ref{method_bioinspired} summarizes the current studies of bio-inspired methodologies.
\begin{table*}[h]   
\centering
\caption{Current Studies of Bio-Inspired Methodologies}
\label{method_bioinspired}   
\resizebox{\linewidth}{!}{
\begin{tabular}{cccccc}    
\toprule
  Reference& Inspiration & Environment& Obstacle avoidance& Coordination&Sensing ability\\    
\midrule  
Tu \textit{et al.}(2021) \cite{tu2021bio}  & Hummingbird & Air (3D) & No & No& Limited\\   
Prasath \textit{et al.}(2022)  \cite{prasath2022dynamics}  &  Ants & Land (2D)& No &Yes &Limited\\   
Lin \textit{et al.}(2011) \cite{lin2011goqbot} & Caterpillar & Land (2D) & No & No& Limited\\  
Nishimura and Mikami(2016) \cite{nishimura2016learning}  & Spider &Land (2D)& No & No &Perfect\\ 
Weymouth \textit{et al.}(2015) \cite{weymouth2015ultra} & Octopus &Underwater (3D)& No &No &Limited\\
He \textit{et al.}(2023) \cite{he2023copebot} & Copepods &Underwater (3D)& No &No &Perfect\\ 
 Berlinger \textit{et al.}(2021) \cite{berlinger2021self}  & Fish & Water surface (2D) &No &Yes& Limited\\
Berlinger \textit{et al.}(2021) \cite{berlinger2021implicit} & Fish & Underwater (3D) &No &Yes& Limited\\
Min and Wang(2011) \cite{min2011design} & Fish & Water surface (2D) &Static &Yes& Perfect\\
Marchese \textit{et al.}(2014) \cite{marchese2014autonomous} & Fish & Underwater (3D) &No &No& Limited\\
\bottomrule   
\end{tabular}  
}
\end{table*}
\begin{figure*}[h]
\centering
\includegraphics[width=1\textwidth]{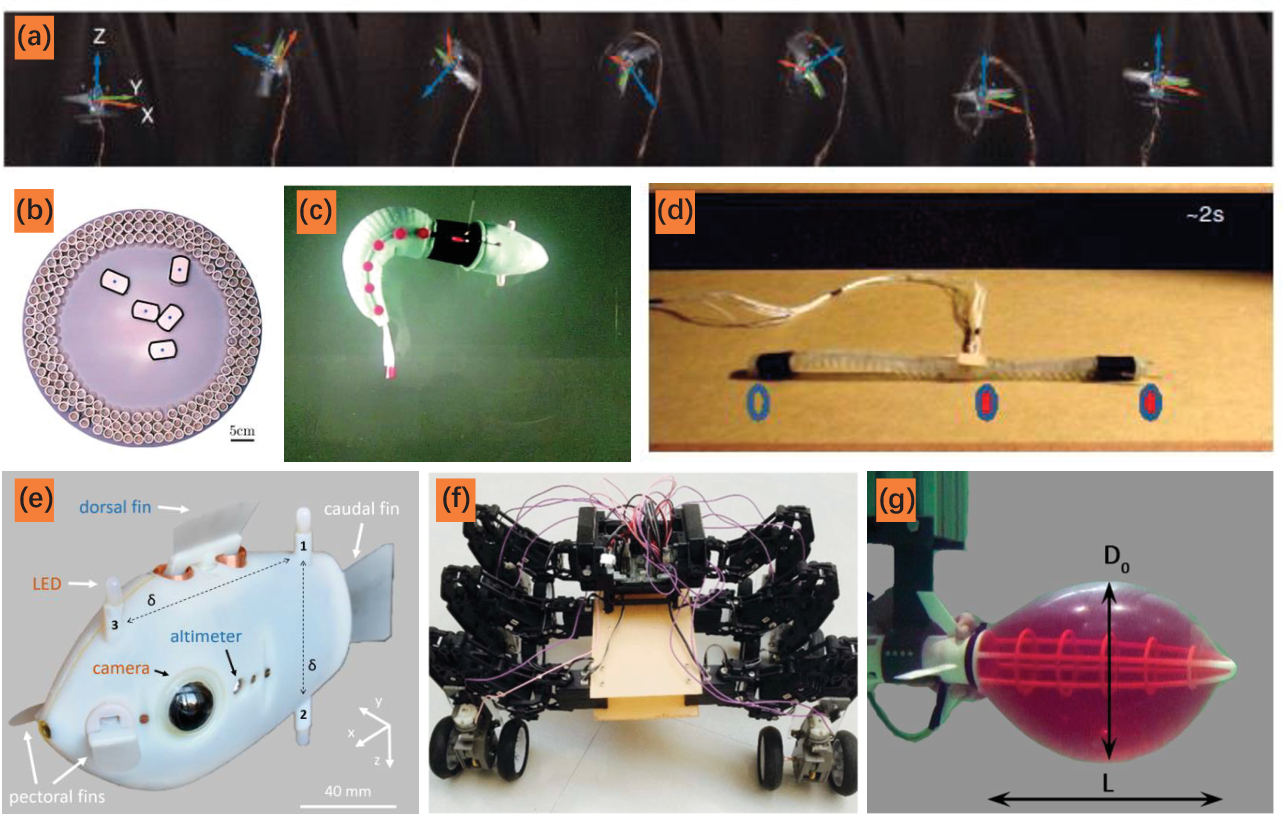}
\caption{Biologically inspired escape robotic systems. (a) a hummingbird-inspired rapid escape with a tight body flips robotic system \cite{tu2021bio}; (b)  multiple robotic systems based on the cooperative excavation and escape of ants \cite{prasath2022dynamics}; (c) a soft fish-inspired robot that is able to change escape direction almost as rapidly as a real-world fish \cite{marchese2014autonomous}; (d) a caterpillar-inspired soft-bodied robot by mimicking the rolling escape behavior of caterpillars \cite{lin2011goqbot}; (e) the self-organized escape of multiple fish-inspired robots based on information from cameras, LEDs, and altimeters \cite{berlinger2021self}; (f) a wheel-legged robot that can autonomously escape from unknown environments \cite{nishimura2016learning}; (g) a flexible hull octopus-inspired robot that is able to rapidly change the size to improve propulsive performance \cite{weymouth2015ultra}. }
\label{fig_bio-isnpired}
\end{figure*} 

There are a variety of different types of animals that are being studied for biologically inspired escape, including insects, fish, birds, and mammals, as shown in Fig. \ref{fig_bio-isnpired}. Each type of animal has its own unique capabilities and limitations, and the choice of animal will depend on the specific needs of the escape scenario.
Papadopoulou \textit{et al.} \cite{papadopoulou2022self} analyzed the GPS trajectories of pigeons in airborne flocks attacked by a robotic falcon to build a species-specific model of collective escape. The model was used to examine a distance-dependent pattern of collective behavior.
Tu \textit{et al.} \cite{tu2021bio} proposed a reinforcement learning to enable animal-like maneuverability on an at-scale, dual-motor actuated flapping wing hummingbird robot. The robot demonstrated a shorter completion time in escape maneuvers compared to the traditional control-based method and successfully performed body flips within one wingspan vertical displacement.
However, the proposed approach still has a gap in real-world applications. The escape behavior of the robot is based on reinforcement learning, which might require a large amount of training data in complex scenarios and might be difficult to handle continuous state spaces.
Prasath \textit{et al.}  \cite{prasath2022dynamics} investigated the collective task of ants escaping from a soft, erodible confining corral. To test the proposed theory over the range of predicted behaviors, they used custom-built robots that respond to stimuli and show the emergence (and failure) of cooperative excavation and escape.
The proposed experiments provide persuasive results of the ant behavior. However, if the proposed experiments could add more complex scenarios, such as obstacles or sudden environmental changes, the results could be more helpful to inspire novel escape algorithms in the field of robotics.
Lin \textit{et al.} \cite{lin2011goqbot} proposed a caterpillar-inspired soft-bodied robot by mimicking the rolling escape behavior of caterpillars. The proposed approach improves the coordination of the robot's body without sensory feedback, providing an estimate of the mechanical power for the robot rolling escape.
The proposed caterpillar-inspired robot is able to navigate in narrow spaces. However, the proposed robot did not consider obstacle avoidance during the escape process.
Nishimura and Mikami \cite{nishimura2016learning} proposed a wheel-legged robot that can autonomously escape from unknown environments using reinforcement learning. The robot uses values of external force measured on the robot's legs as the definition of states and rewards to reduce the number of states and actions.
However, the escape control of the wheel-legged robot is based on reinforcement learning which might need a lot of experimentation and testing may be required to collect data.
Weymouth \textit{et al.} \cite{weymouth2015ultra} designed and tested a flexible hull octopus-inspired robot, which demonstrates exceptional fast-starting performance. The rapid change in the size of the robot helps reduce separation in bluff bodies, recovers fluid energy, and improves propulsive performance. 
However, the proposed experiments did not consider the obstacle and potential environmental changes.
He \textit{et al.} \cite{he2023copebot} mimicked the escape behavior of copepods, which allows robots with high movability to escape the potential threat.

\begin{figure}[h]
\centering
\includegraphics[width=0.35\textwidth]{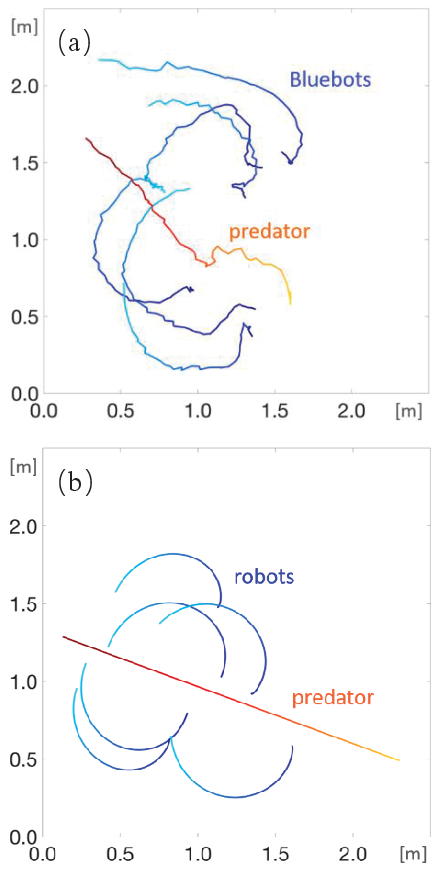}
\caption{Escape trajectories of fish-inspired robots \cite{berlinger2021self}. (a) experiment with six Bluebots. (b) simulation with six robots.}
\label{Fish_escape}
\end{figure}

Fishes are well-known examples that possess the ability to navigate and respond effectively to dynamic environments through the use of simple mechanisms. Through cooperation and limited implicit communication, fish schools are able to accomplish complex tasks that would be beyond the capabilities of an individual fish \cite{doran2022fish}. 
The main sensory modalities of fish to achieve group escape are visual observations between individuals, and the detecting of water movements based on their mechanosensory lateral line \cite{ioannou2011social}.  The fountain maneuver model is able to reproduce many collective features in fish schools \cite{ishiwaka2021foids}. Thus, the fountain maneuver model was copied into the robotic system, providing a solution to control robots with distributed control architecture.
Berlinger \textit{et al.} \cite{berlinger2021self} used the fountain maneuver model to design the escape group of the underwater robotic platform. As shown in Fig. \ref{Fish_escape}, the simulation and experimental results show that robots can escape the predator while keeping a constantly visible angle to the predator. However, their study only considers the collision-free environment. 
The artificial virtual forces from nearby neighbors based on relative distance are another method of achieving the collective behaviors of swarm robots. Berlinger \textit{et al.} \cite{berlinger2021implicit} used a potential field-based model to mimic collective behaviors for underwater robots in 3D environments. Min and Wang \cite{min2011design} proposed a fish-inspired escape algorithm based on Newton-Euler dynamics equations, which is capable of rapidly escaping the predator and avoiding collision with obstacles. However, their study only considers the static environment, where the obstacle and the threat are continuously stable in the environment. Marchese \textit{et al.} \cite{marchese2014autonomous} proposed a soft fish-inspired robot that is able to change escape direction almost as rapidly as a real-world fish.

As mentioned in the literature reviews, the current state of bio-inspired methodologies is analyzed as follows:
 \begin{enumerate}
\item  Most studies did not consider obstacle avoidance (e.g.,\cite{tu2021bio,prasath2022dynamics,lin2011goqbot,nishimura2016learning,weymouth2015ultra,he2023copebot,berlinger2021self,berlinger2021implicit,marchese2014autonomous}). This may imply that bio-inspired methodologies are limited focus on obstacle avoidance. Further research could explore the integration of obstacle-avoidance capabilities into these methodologies.
\item Some researchers considered the coordination issue (e.g.,\cite{prasath2022dynamics,berlinger2021self,berlinger2021implicit,min2011design}), indicating that bio-inspired methodologies often emphasize cooperative behavior, inspired by natural swarm intelligence and collective behavior.
\item  Most studies considered limited sensing capabilities, which are more reflective of real-world scenarios (e.g.,\cite{tu2021bio,prasath2022dynamics,lin2011goqbot,weymouth2015ultra,berlinger2021self,berlinger2021implicit,marchese2014autonomous}). This suggests that bio-inspired methodologies may be better suited for practical applications, as they account for the inherent limitations and uncertainties in robotic sensing systems.
\end{enumerate}
In summary, bio-inspired methodologies in robotics draw from diverse natural inspirations and are applicable across various environments. Although obstacle avoidance has not been a primary focus in these studies, multi-robot coordination is frequently investigated. The consideration of limited sensing capabilities in most studies indicates that bio-inspired methodologies may be more adaptable and relevant to real-world applications.

\subsection{Comparison Analysis of the Current Research}
Planning-based methodologies are able to plan efficient and safe paths considering obstacle situations. Compared to other methodologies, one of the primary advantages of planning-based methodologies is their ability to avoid obstacles and deal with other environmental factors (e.g. \cite{gilles2020evasive, hu2019multiobjective, wang2021path, saravanakumar2021sampling, dong2012strategies, chen2020dynamic, ajeil2021novel, zaccone2020collision, tang2021gwo, goodwin2015escape, zhu2020receding}). In real-world scenarios, there may be a range of obstacles that need to be navigated to reach safety.  Planning-based methodologies are able to generate safe routes through these obstacles, allowing robots to reach safety quickly and without harm. Another advantage of planning-based methodologies is their ability to adapt to changing circumstances (e.g. \cite{hu2019multiobjective,chen2020dynamic, ajeil2021novel, zaccone2020collision,zhu2020receding}).  Planning-based methodologies are able to take these changes into account and adjust the escape path accordingly. Compared to partitioning-based and learning-based methodologies, most planning-based methodologies require only small-scale experimentation and testing to achieve escape planning.

Partitioning-based methodologies have been widely used in pursuer-evasion games. Compared to other methodologies, one of the critical advantages of partitioning-based methodologies is their ability to provide global solutions to the evasive strategy with full information (e.g.\cite{liang2023collaborative, ramana2017pursuit, sun2022cooperative, kumar2019experimental, oyler2016pursuit, yan2018reach}). Another advantage is that partitioning-based methodologies are able to provide a comparative result for the influence of the obstacle. In other methodologies, the influence of the obstacle is only considered as the obstacle-avoidance problem. However, the obstacle could create some advantages for the evader robot to improve the escape performance, such as blocking the possible pursuit path from the pursuer. In partitioning-based methodologies, the influence of the obstacle can be analyzed by partitioning different areas to improve the escape performance (e.g.,\cite{liang2023collaborative, sun2022cooperative, oyler2016pursuit}).

Learning-based methodologies are particularly useful in applications where the desired outcome is not predetermined and the robot has to learn through trial and error. Compared to other methodologies, learning-based methodologies offer a valuable advantage in avoiding complex differential equations for escape scenarios, such as Equation \ref{diff1}-\ref{diff3}. By establishing the environment, learning-based methodologies enable the direct acquisition of the escape strategy (e.g.\cite{di2019optimizing,wang2019fuzzy, weitzenfeld2008prey,qi2020deep,zhang2022game,xu2022autonomous,xu2022pursuit,ji2021obstacle,gao2019fast}). Another advantage is that learning-based methodologies enable robots to acquire complex behaviors through interactions with an unknown environment, without prior knowledge. The robot receives rewards from the environment and leverages its existing knowledge to enhance its future performance. Therefore, learning-based methodologies offer adaptability to diverse environments and escape strategies (e.g.,\cite{di2019optimizing,wang2019fuzzy, weitzenfeld2008prey,qi2020deep,zhang2022game,xu2022autonomous,xu2022pursuit,ji2021obstacle,gao2019fast,shree2022learning}).

The bio-inspired methodologies have been applied to a number of escape tasks and practical applications. Compared to other methodologies, one advantage of bio-inspired methodologies is the scalability and flexibility of bio-inspired robotic systems, as individual robots are able to add or delete dynamically, without requiring any explicit reorganization (e.g.,\cite{berlinger2021self, berlinger2021implicit, min2011design, marchese2014autonomous}). 
Another advantage is that the autonomy and self-sufficiency of the robot allow it to adapt quickly to rapidly changing situations. The use of bio-inspired robots could result in a relatively less costly system, which could improve robust of the robotic systems in the sense that robotic systems are less prone to experiencing failures (e.g. \cite{tu2021bio, prasath2022dynamics, lin2011goqbot, nishimura2016learning, weymouth2015ultra, he2023copebot}).

\section{Applications of Intelligent Escape}
\label{sec:application}
Intelligent robotic escape has potential applications in various domains such as rescue and evacuation, military security, and healthcare service. However, designing real-world applications of intelligent robotic escape requires addressing multiple objectives. These objectives may conflict with each other and entail tradeoffs in terms of costs and benefits. In this section, previous and potential applications are summarized to provide insight for future development.
\begin{figure*}[h]
\centering
\includegraphics[width=0.9\textwidth]{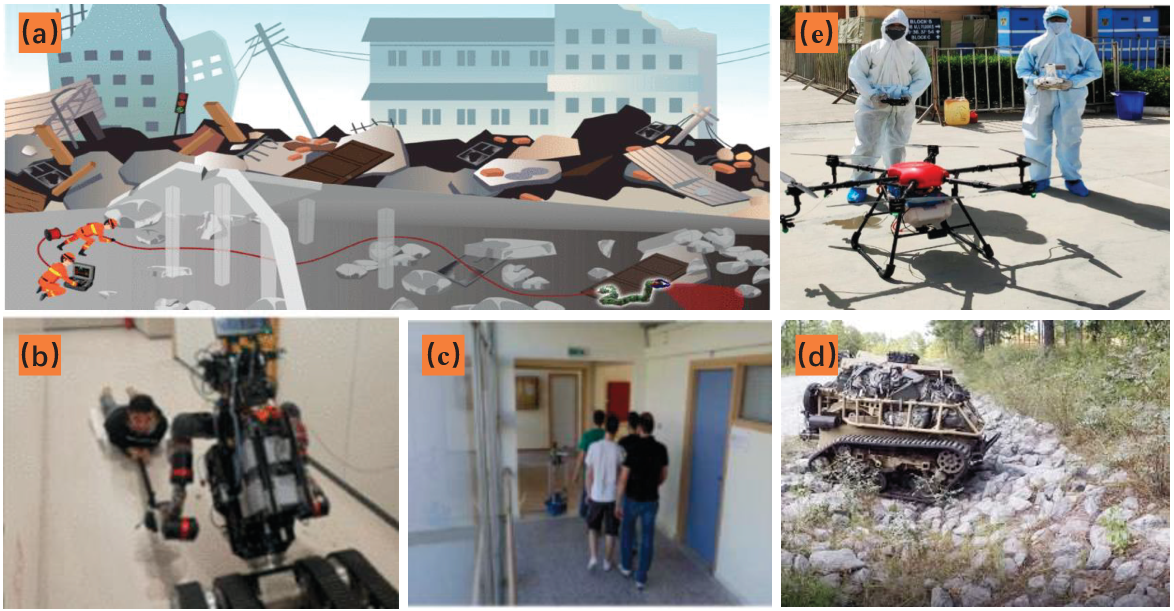}
\caption{Potential applications of intelligent escape. (a) search and rescue tasks in narrow spaces \cite{han2022snake}; (b) humanoid robot to rescue the target in complex environments \cite{sun2021bit}; (c) robot-assisted evacuation in indoor environments \cite{boukas2014robot}; (d) military robots in field environments \cite{lopez2018new}; (e) aerial robots to protect the community during COVID-19 pandemic \cite{kumar2021drone}. }
\label{fig_applications}
\end{figure*} 
\subsection{Search and Rescue}
The application of robot escape in search and rescue operations has shown significant potential in improving the efficiency and safety of rescue workers while saving more lives. This technique involves using robots to navigate through challenging terrain, debris, and obstacles to locate and extract survivors from hazardous environments. One significant advantage of using robot escape in search and rescue is the increased efficiency of the rescue operation \cite{han2022snake}. Robots can work tirelessly without the risk of exhaustion or injury, allowing them to cover more ground and locate survivors quickly. In addition, intelligent escape is able to improve  the safety of rescue workers \cite{sun2021bit}. Hazardous environments such as collapsed buildings, mines, or natural disaster sites pose significant risks to rescue workers, including the risk of injury or death. 

The robotic systems are designed to navigate challenging environments to detect, locate, and assist victims in emergency scenarios, improving response efficiency and effectiveness \cite{drew2021multi}.  
Wang \textit{et al.}  \cite{wang2020three} developed an improved visual SLAM algorithm specifically designed for search and rescue scenarios, which allows efficient and accurate mapping of complex and dynamic environments.
Dong\textit{et al.}  \cite{dong2021uav} proposed a novel algorithm that combines computer vision and machine learning techniques to efficiently detect and locate survivors in disaster-stricken areas. 
The intelligent escape can empower the robotic systems to adapt and respond effectively to new emergency situations or disasters during the search and rescue process.

\subsection{Evacuation}
The application of robot escape in evacuation refers to the use of robots to assist in the evacuation process during emergency situations such as natural disasters, fires, or terrorist attacks. The purpose of this discussion is to explore the benefits and limitations of the application of robot escape in evacuation.



Robot-assisted evacuation is a growing area of research that focuses on the use of robots to aid in the evacuation of people from buildings and other structures during emergency situations \cite{zhou2022robot}. The role of mobile robots is similar to that of a trained leader in guided crowd evacuation and can provide route information to evacuees and/or lead occupants to safety. In some accidents, such as nuclear leakage, a trained leader cannot be assigned to guide evacuees out for security reasons \cite{zhou2019guided}. Sakour \textit{et al.} \cite{sakour2017robot} proposed a review of robot-assisted crowd evacuation under emergency situations. Tang \textit{et al.} \cite{tang2016human} proposed a robot-guided evacuation system using a dynamic environment information algorithm and panic model simulations to improve evacuation efficiency in complex indoor environments. Jiang \textit{et al.} \cite{jiang2017learning}  proposes a robot-assisted pedestrian regulation system to regulate flows during emergency evacuation and investigates passive human-robot interaction. The research on crowd evacuation with mobile robots was also performed from the perspective of trust in human-robot interactions \cite{boukas2014robot}.

\subsection{Military Security}
One of the primary benefits of using robots to escape dangerous situations is that they can reduce the risk of human casualties. For example, if soldiers are trapped in a building with no escape route, sending in a robot to locate an escape route can potentially save lives. Additionally, robots can be designed to withstand harsh environmental conditions, such as extreme heat or cold, making them ideal for use in situations where human survival may be difficult. Another benefit of robot escape is that it can provide valuable reconnaissance information to military personnel. By sending a robot into a dangerous situation first, soldiers can gather critical intelligence about the environment and potential threats before entering themselves. 
For instance, the RS2-H1 system can help to reduce the risk of surprise attacks and improve the overall safety of military operations \cite{lopez2018new}.
Intelligent escape in military applications is the use of robots to avoid detection, injury, or capture by enemy forces. Robotic platforms can provide remote surgical support and evacuation activities in war zones \cite{panesar2019artificial}. Autonomous drones that can fly stealthily and evade enemy radars. Swarm robotics that can coordinate with each other and adapt to changing environments \cite{wang2020use}. Robot evasion in military applications can have advantages such as enhancing the safety, efficiency, and accuracy of military operations. However, it can also pose risks such as malfunctions, glitches, ethical dilemmas, and loss of human control.

\subsection{Healthcare}

Robots are increasingly being used in healthcare settings to assist with tasks such as patient lifting, medication delivery, and surgical procedures \cite{kumar2021drone}. In these settings, it is important for robots to operate within designated areas and to avoid collisions with people or equipment. Effective methods to prevent robot escapes can help ensure that robots can perform their tasks safely and accurately while minimizing the risk of injury or error. Robots can undertake human-like activities and can be gainfully programmed to replace some of the human interactions \cite{javaid2020robotics}. Robots are available for both natural disasters and the latest biological nemesis \cite{javaid2020industry}. They are helpful for clinical care such as screening, diagnosis, and disease prevention. Robotics applications are applied to address the risks of infectious diseases and have the capability to handle any epidemic or pandemic in the future.

\section{Challenges and Future Trends}
\label{sec:challenges}
The future of intelligent robot escape hinges on addressing key challenges and embracing emerging trends. By tackling these challenges, robotic systems improve the adaptability, resilience, and efficiency of robotic systems in dynamic and complex environments.

\subsection{Real-Time Collision-Free Escape in Complex Scenarios}
As intelligent robotics continues to advance, researchers are striving to enhance the adaptability and responsiveness of these systems in complex and dynamic environments. 
As shown in Tables \ref{method_pathplanning}-\ref{method_bioinspired}, most current methodologies do not consider obstacle avoidance, while some studies investigate static obstacle avoidance. However, obstacle avoidance is a critical capability of robotic systems for real-world applications, particularly in dynamic and rapidly changing environments. Kyprianou \textit{et al.} 
 proposed a comprehensive survey of the achievement and importance 
of multi-robot systems in dynamic environments \cite{kyprianou2022towards}. 
Kobayashi and Motoi pointed out that  mobile robots considering static and dynamic obstacles is one of the challenging research topics \cite{kobayashi2022local}. Huang \textit{et al.} analyzed the collision-avoidance ability of robots is able to reduce traffic accidents and relieve traffic congestion in real-world applications \cite{huang2022sine}. Therefore, a 
pivotal aspect of this endeavor lies in the development of real-time collision-free escape mechanisms, which enable robots to swiftly and safely navigate through intricate spaces while avoiding potential obstacles. Using state-of-the-art algorithms, sensor fusion techniques, and machine learning methodologies, researchers aim to create sophisticated models that can accurately predict and respond to environmental changes, thereby facilitating prompt and efficient escape maneuvers. This advancement in real-time collision-free escape not only holds the promise of significantly improving the performance and reliability of autonomous robotic systems, but also has far-reaching implications for a wide array of applications, including search and rescue, surveillance, and disaster response. Ultimately, this line of research will contribute to the emergence of intelligent robotic systems that can seamlessly integrate into and navigate through the ever-changing complexities of the real world.

\subsection{Robust Escape Considering Uncertainties and Disturbances}
In the realm of intelligent robotics, the ability to effectively manage and respond to uncertainties and disturbances is a crucial factor in determining the success of escape strategies. To address this challenge, future research efforts should focus on developing robust and adaptive escape mechanisms that can accommodate the inherent unpredictability of real-world environments. 
Fareh \textit{et al.} analyzed the active disturbance rejection control structure to solve robotic application problems \cite{fareh2021active}. In addition, aerial and underwater environments may impose uncertainties and disturbances to robot systems. For instance, Liang \textit{et al.}  proposed a trajectory tracking control method under model uncertainty and unknown disturbances in an aerial environment \cite{liang2021low}. Chu \textit{et al.} proposed a path planning method for underwater robots under ocean current disturbance \cite{chu2022path}.

These mechanisms should incorporate advanced probabilistic models, stochastic control algorithms, and resilient system architectures to enable robots to continuously adapt to changing conditions and maintain optimal performance in the face of uncertainty. By incorporating such techniques, robotic systems will be able to successfully navigate and escape complex scenarios involving ambiguous sensory data, unforeseen obstacles, and dynamic environmental changes. The development of escape strategies that can effectively handle uncertainties and disturbances will not only enhance the overall resilience and reliability of intelligent robotic systems but also lay the groundwork for a new generation of autonomous machines capable of operating in an increasingly unpredictable world.

 \subsection{Cooperative Intelligence for Collaborative Escape Solutions}
As the complexity of tasks and environments that intelligent robots are expected to navigate increases, a growing emphasis is being placed on the development of cooperative escape strategies. In contrast to traditional, individual-based approaches, intelligent cooperation involves the coordination and collaboration of multiple robotic agents, allowing them to pool their resources, knowledge, and capabilities in order to enhance their collective ability to adapt and respond to challenging situations. 
For example, a team of robots equipped with different sensors can provide more comprehensive and accurate situational awareness than a single robot \cite{ferrer2021secure}. By sharing their information and collaborating on their actions, the robots can work together to navigate complex environments, avoid obstacles, and locate escape routes more efficiently and effectively \cite{yang2021review,simetti2021wimust,zhou2021survey}.
In addition, harnessing advanced communication protocols, distributed control algorithms, and swarm intelligence principles, cooperative escape strategies enable robotic teams to share information, make joint decisions, and dynamically reconfigure their roles and formations, resulting in more efficient and effective escape maneuvers. By fostering synergistic interactions among robots, the development of cooperative intelligent escape strategies not only bolsters the overall performance and robustness of robotic systems but also paves the way for the emergence of complex, decentralized networks of autonomous machines that can seamlessly collaborate to achieve their objectives in dynamic and uncertain environments.

\subsection{Innovating Escape Mechanisms with Bio-Inspired Approaches}
In light of the rapid advancements in intelligent robotics witnessed over the past decades, the field has now reached a critical juncture where researchers must devise innovative strategies to address the emergent challenges in robot escape mechanisms. One particularly promising avenue for the future development of intelligent robot escape is the exploration of bio-inspired escape approaches, which draw upon the remarkable adaptability and resilience found in biological organisms. By integrating principles from disciplines such as evolutionary biology, ethology, and biomechanics, these approaches have the potential to significantly enhance the robustness, versatility, and energy efficiency of robotic systems, enabling them to better navigate and adapt to complex, unpredictable environments. 
For instance, Huang \textit{et al.} proposed a co-evolutionary algorithm to design the high accuracy evasive strategy \cite{huang2021novel}. Agarwal and Bharti proposed a swarm intelligence algorithm to achieve high-efficiency obstacle avoidance \cite{agarwal2021implementing}. Moorthy and Joo proposed a neurodynamics-based method for distributed leader-following formation control \cite{moorthy2022distributed}.
The interdisciplinary synergy would not only usher in a new era of intelligent robot design, but also contribute to a deeper understanding of the underlying mechanisms governing natural escape behaviors.

\section{Conclusion}
This paper is a comprehensive survey of current research on the intelligent escape of robotic systems. The survey focuses on four critical methods of intelligent escape, which include planning-based methodologies, partitioning-based methodologies, learning-based methodologies, and bio-inspired methodologies. In addition, the strengths and limitations of each method are analyzed, and potential applications are discussed. The insights provided in this survey could help guide future research in developing more effective and adaptive escape behaviors for robotic systems.
\label{sec:conclusion}


\bibliography{sn-bibliography}


\begin{thebibliography}{95}
\ifx \bisbn   \undefined \def \bisbn  #1{ISBN #1}\fi
\ifx \binits  \undefined \def \binits#1{#1}\fi
\ifx \bauthor  \undefined \def \bauthor#1{#1}\fi
\ifx \batitle  \undefined \def \batitle#1{#1}\fi
\ifx \bjtitle  \undefined \def \bjtitle#1{#1}\fi
\ifx \bvolume  \undefined \def \bvolume#1{\textbf{#1}}\fi
\ifx \byear  \undefined \def \byear#1{#1}\fi
\ifx \bissue  \undefined \def \bissue#1{#1}\fi
\ifx \bfpage  \undefined \def \bfpage#1{#1}\fi
\ifx \blpage  \undefined \def \blpage #1{#1}\fi
\ifx \burl  \undefined \def \burl#1{\textsf{#1}}\fi
\ifx \doiurl  \undefined \def \doiurl#1{\url{https://doi.org/#1}}\fi
\ifx \betal  \undefined \def \betal{\textit{et al.}}\fi
\ifx \binstitute  \undefined \def \binstitute#1{#1}\fi
\ifx \binstitutionaled  \undefined \def \binstitutionaled#1{#1}\fi
\ifx \bctitle  \undefined \def \bctitle#1{#1}\fi
\ifx \beditor  \undefined \def \beditor#1{#1}\fi
\ifx \bpublisher  \undefined \def \bpublisher#1{#1}\fi
\ifx \bbtitle  \undefined \def \bbtitle#1{#1}\fi
\ifx \bedition  \undefined \def \bedition#1{#1}\fi
\ifx \bseriesno  \undefined \def \bseriesno#1{#1}\fi
\ifx \blocation  \undefined \def \blocation#1{#1}\fi
\ifx \bsertitle  \undefined \def \bsertitle#1{#1}\fi
\ifx \bsnm \undefined \def \bsnm#1{#1}\fi
\ifx \bsuffix \undefined \def \bsuffix#1{#1}\fi
\ifx \bparticle \undefined \def \bparticle#1{#1}\fi
\ifx \barticle \undefined \def \barticle#1{#1}\fi
\bibcommenthead
\ifx \bconfdate \undefined \def \bconfdate #1{#1}\fi
\ifx \botherref \undefined \def \botherref #1{#1}\fi
\ifx \url \undefined \def \url#1{\textsf{#1}}\fi
\ifx \bchapter \undefined \def \bchapter#1{#1}\fi
\ifx \bbook \undefined \def \bbook#1{#1}\fi
\ifx \bcomment \undefined \def \bcomment#1{#1}\fi
\ifx \oauthor \undefined \def \oauthor#1{#1}\fi
\ifx \citeauthoryear \undefined \def \citeauthoryear#1{#1}\fi
\ifx \endbibitem  \undefined \def \endbibitem {}\fi
\ifx \bconflocation  \undefined \def \bconflocation#1{#1}\fi
\ifx \arxivurl  \undefined \def \arxivurl#1{\textsf{#1}}\fi
\csname PreBibitemsHook\endcsname

\bibitem{casini2022two}
\begin{barticle}
\bauthor{\bsnm{Casini}, \binits{M.}},
\bauthor{\bsnm{Garulli}, \binits{A.}}:
\batitle{A two-pursuer one-evader game with equal speed and finite capture radius}.
\bjtitle{Journal of Intelligent \& Robotic Systems}
\bvolume{106}(\bissue{4}),
\bfpage{77}
(\byear{2022})
\end{barticle}
\endbibitem

\bibitem{lopez2019solutions}
\begin{barticle}
\bauthor{\bsnm{Lopez}, \binits{V.G.}},
\bauthor{\bsnm{Lewis}, \binits{F.L.}},
\bauthor{\bsnm{Wan}, \binits{Y.}},
\bauthor{\bsnm{Sanchez}, \binits{E.N.}},
\bauthor{\bsnm{Fan}, \binits{L.}}:
\batitle{Solutions for multiagent pursuit-evasion games on communication graphs: Finite-time capture and asymptotic behaviors}.
\bjtitle{IEEE Transactions on Automatic Control}
\bvolume{65}(\bissue{5}),
\bfpage{1911}--\blpage{1923}
(\byear{2019})
\end{barticle}
\endbibitem

\bibitem{xu2022multiplayer}
\begin{barticle}
\bauthor{\bsnm{Xu}, \binits{Y.}},
\bauthor{\bsnm{Yang}, \binits{H.}},
\bauthor{\bsnm{Jiang}, \binits{B.}},
\bauthor{\bsnm{Polycarpou}, \binits{M.M.}}:
\batitle{Multiplayer pursuit-evasion differential games with malicious pursuers}.
\bjtitle{IEEE Transactions on Automatic Control}
\bvolume{67}(\bissue{9}),
\bfpage{4939}--\blpage{4946}
(\byear{2022})
\end{barticle}
\endbibitem

\bibitem{yasin2020unmanned}
\begin{barticle}
\bauthor{\bsnm{Yasin}, \binits{J.N.}},
\bauthor{\bsnm{Mohamed}, \binits{S.A.}},
\bauthor{\bsnm{Haghbayan}, \binits{M.-H.}},
\bauthor{\bsnm{Heikkonen}, \binits{J.}},
\bauthor{\bsnm{Tenhunen}, \binits{H.}},
\bauthor{\bsnm{Plosila}, \binits{J.}}:
\batitle{Unmanned aerial vehicles (uavs): Collision avoidance systems and approaches}.
\bjtitle{IEEE access}
\bvolume{8},
\bfpage{105139}--\blpage{105155}
(\byear{2020})
\end{barticle}
\endbibitem

\bibitem{zhang2020mutual}
\begin{barticle}
\bauthor{\bsnm{Zhang}, \binits{Z.}},
\bauthor{\bsnm{Zheng}, \binits{L.}},
\bauthor{\bsnm{Chen}, \binits{Z.}},
\bauthor{\bsnm{Kong}, \binits{L.}},
\bauthor{\bsnm{Karimi}, \binits{H.R.}}:
\batitle{Mutual-collision-avoidance scheme synthesized by neural networks for dual redundant robot manipulators executing cooperative tasks}.
\bjtitle{IEEE transactions on neural networks and learning systems}
\bvolume{32}(\bissue{3}),
\bfpage{1052}--\blpage{1066}
(\byear{2020})
\end{barticle}
\endbibitem

\bibitem{ding2020collision}
\begin{bchapter}
\bauthor{\bsnm{Ding}, \binits{Y.}},
\bauthor{\bsnm{Thomas}, \binits{U.}}:
\bctitle{Collision avoidance with proximity servoing for redundant serial robot manipulators}.
In: \bbtitle{2020 IEEE International Conference on Robotics and Automation (ICRA)},
pp. \bfpage{10249}--\blpage{10255}
(\byear{2020}).
\bcomment{IEEE}
\end{bchapter}
\endbibitem

\bibitem{li2016dynamics}
\begin{barticle}
\bauthor{\bsnm{Li}, \binits{W.}}:
\batitle{A dynamics perspective of pursuit-evasion: capturing and escaping when the pursuer runs faster than the agile evader}.
\bjtitle{IEEE Transactions on Automatic Control}
\bvolume{62}(\bissue{1}),
\bfpage{451}--\blpage{457}
(\byear{2016})
\end{barticle}
\endbibitem

\bibitem{currier2020bio}
\begin{barticle}
\bauthor{\bsnm{Currier}, \binits{T.M.}},
\bauthor{\bsnm{Lheron}, \binits{S.}},
\bauthor{\bsnm{Modarres-Sadeghi}, \binits{Y.}}:
\batitle{A bio-inspired robotic fish utilizes the snap-through buckling of its spine to generate accelerations of more than 20g}.
\bjtitle{Bioinspiration \& Biomimetics}
\bvolume{15}(\bissue{5}),
\bfpage{055006}
(\byear{2020})
\end{barticle}
\endbibitem

\bibitem{boyer2021towards}
\begin{barticle}
\bauthor{\bsnm{Boyer}, \binits{A.}},
\bauthor{\bsnm{Farzaneh}, \binits{F.}}:
\batitle{Towards an ethic of robotics}.
\bjtitle{Journal of Organizational Psychology}
\bvolume{21}(\bissue{3}),
\bfpage{84}--\blpage{100}
(\byear{2021})
\end{barticle}
\endbibitem

\bibitem{birajdar2020development}
\begin{barticle}
\bauthor{\bsnm{Birajdar}, \binits{G.S.}},
\bauthor{\bsnm{Singh}, \binits{R.}},
\bauthor{\bsnm{Gehlot}, \binits{A.}},
\bauthor{\bsnm{Thakur}, \binits{A.K.}}:
\batitle{Development in building fire detection and evacuation system-a comprehensive review}.
\bjtitle{Int. J. Electr. Comput. Eng.(IJECE)}
\bvolume{10},
\bfpage{6644}--\blpage{6654}
(\byear{2020})
\end{barticle}
\endbibitem

\bibitem{muhammad2019designing}
\begin{barticle}
\bauthor{\bsnm{Muhammad}, \binits{S.}},
\bauthor{\bsnm{Mohammad}, \binits{N.}},
\bauthor{\bsnm{Bashar}, \binits{A.}},
\bauthor{\bsnm{Khan}, \binits{M.A.}}:
\batitle{Designing human assisted wireless sensor and robot networks using probabilistic model checking}.
\bjtitle{Journal of Intelligent \& Robotic Systems}
\bvolume{94},
\bfpage{687}--\blpage{709}
(\byear{2019})
\end{barticle}
\endbibitem

\bibitem{haddeler2020evaluation}
\begin{barticle}
\bauthor{\bsnm{Haddeler}, \binits{G.}},
\bauthor{\bsnm{Aybakan}, \binits{A.}},
\bauthor{\bsnm{Akay}, \binits{M.C.}},
\bauthor{\bsnm{Temeltas}, \binits{H.}}:
\batitle{Evaluation of 3d lidar sensor setup for heterogeneous robot team}.
\bjtitle{Journal of Intelligent \& Robotic Systems}
\bvolume{100},
\bfpage{689}--\blpage{709}
(\byear{2020})
\end{barticle}
\endbibitem

\bibitem{jin2021bioinspired}
\begin{barticle}
\bauthor{\bsnm{Jin}, \binits{G.}},
\bauthor{\bsnm{Sun}, \binits{Y.}},
\bauthor{\bsnm{Geng}, \binits{J.}},
\bauthor{\bsnm{Yuan}, \binits{X.}},
\bauthor{\bsnm{Chen}, \binits{T.}},
\bauthor{\bsnm{Liu}, \binits{H.}},
\bauthor{\bsnm{Wang}, \binits{F.}},
\bauthor{\bsnm{Sun}, \binits{L.}}:
\batitle{Bioinspired soft caterpillar robot with ultra-stretchable bionic sensors based on functional liquid metal}.
\bjtitle{Nano Energy}
\bvolume{84},
\bfpage{105896}
(\byear{2021})
\end{barticle}
\endbibitem

\bibitem{gilles2020evasive}
\begin{barticle}
\bauthor{\bsnm{Gilles}, \binits{M.A.}},
\bauthor{\bsnm{Vladimirsky}, \binits{A.}}:
\batitle{Evasive path planning under surveillance uncertainty}.
\bjtitle{Dynamic Games and Applications}
\bvolume{10}(\bissue{2}),
\bfpage{391}--\blpage{416}
(\byear{2020})
\end{barticle}
\endbibitem

\bibitem{gong2022online}
\begin{barticle}
\bauthor{\bsnm{Gong}, \binits{Z.}},
\bauthor{\bsnm{He}, \binits{B.}},
\bauthor{\bsnm{Hu}, \binits{C.}},
\bauthor{\bsnm{Zhang}, \binits{X.}},
\bauthor{\bsnm{Kang}, \binits{W.}}:
\batitle{Online adaptive dynamic programming-based solution of networked multiple-pursuer and single-evader game}.
\bjtitle{Electronics}
\bvolume{11}(\bissue{21}),
\bfpage{3583}
(\byear{2022})
\end{barticle}
\endbibitem

\bibitem{zhou2022decentralized}
\begin{barticle}
\bauthor{\bsnm{Zhou}, \binits{Z.}},
\bauthor{\bsnm{Xu}, \binits{H.}}:
\batitle{Decentralized optimal large scale multi-player pursuit-evasion strategies: A mean field game approach with reinforcement learning}.
\bjtitle{Neurocomputing}
\bvolume{484},
\bfpage{46}--\blpage{58}
(\byear{2022})
\end{barticle}
\endbibitem

\bibitem{scott2018optimal}
\begin{barticle}
\bauthor{\bsnm{Scott}, \binits{W.L.}},
\bauthor{\bsnm{Leonard}, \binits{N.E.}}:
\batitle{Optimal evasive strategies for multiple interacting agents with motion constraints}.
\bjtitle{Automatica}
\bvolume{94},
\bfpage{26}--\blpage{34}
(\byear{2018})
\end{barticle}
\endbibitem

\bibitem{lewis2018minimum}
\begin{botherref}
\oauthor{\bsnm{Lewis~Scott}, \binits{W.}},
\oauthor{\bsnm{Ehrich~Leonard}, \binits{N.}}:
Minimum-time trajectories for steered agent with constraints on speed, lateral acceleration, and turning rate.
Journal of Dynamic Systems, Measurement, and Control
\textbf{140}(7)
(2018)
\end{botherref}
\endbibitem

\bibitem{makkapati2019optimal}
\begin{barticle}
\bauthor{\bsnm{Makkapati}, \binits{V.R.}},
\bauthor{\bsnm{Tsiotras}, \binits{P.}}:
\batitle{Optimal evading strategies and task allocation in multi-player pursuit--evasion problems}.
\bjtitle{Dynamic Games and Applications}
\bvolume{9},
\bfpage{1168}--\blpage{1187}
(\byear{2019})
\end{barticle}
\endbibitem

\bibitem{hu2019multiobjective}
\begin{barticle}
\bauthor{\bsnm{Hu}, \binits{L.}},
\bauthor{\bsnm{Naeem}, \binits{W.}},
\bauthor{\bsnm{Rajabally}, \binits{E.}},
\bauthor{\bsnm{Watson}, \binits{G.}},
\bauthor{\bsnm{Mills}, \binits{T.}},
\bauthor{\bsnm{Bhuiyan}, \binits{Z.}},
\bauthor{\bsnm{Raeburn}, \binits{C.}},
\bauthor{\bsnm{Salter}, \binits{I.}},
\bauthor{\bsnm{Pekcan}, \binits{C.}}:
\batitle{A multiobjective optimization approach for colregs-compliant path planning of autonomous surface vehicles verified on networked bridge simulators}.
\bjtitle{IEEE Transactions on Intelligent Transportation Systems}
\bvolume{21}(\bissue{3}),
\bfpage{1167}--\blpage{1179}
(\byear{2019})
\end{barticle}
\endbibitem

\bibitem{wang2021path}
\begin{barticle}
\bauthor{\bsnm{Wang}, \binits{J.}},
\bauthor{\bsnm{Yan}, \binits{Y.}},
\bauthor{\bsnm{Zhang}, \binits{K.}},
\bauthor{\bsnm{Chen}, \binits{Y.}},
\bauthor{\bsnm{Cao}, \binits{M.}},
\bauthor{\bsnm{Yin}, \binits{G.}}:
\batitle{Path planning on large curvature roads using driver-vehicle-road system based on the kinematic vehicle model}.
\bjtitle{IEEE Transactions on Vehicular Technology}
\bvolume{71}(\bissue{1}),
\bfpage{311}--\blpage{325}
(\byear{2021})
\end{barticle}
\endbibitem

\bibitem{saravanakumar2021sampling}
\begin{barticle}
\bauthor{\bsnm{Saravanakumar}, \binits{A.}},
\bauthor{\bsnm{Kaviyarasu}, \binits{A.}},
\bauthor{\bsnm{Ashly~Jasmine}, \binits{R.}}:
\batitle{Sampling based path planning algorithm for uav collision avoidance}.
\bjtitle{S{\=a}dhan{\=a}}
\bvolume{46}(\bissue{3}),
\bfpage{112}
(\byear{2021})
\end{barticle}
\endbibitem

\bibitem{dong2012strategies}
\begin{bchapter}
\bauthor{\bsnm{Dong}, \binits{J.}},
\bauthor{\bsnm{Zhang}, \binits{X.}},
\bauthor{\bsnm{Jia}, \binits{X.}}:
\bctitle{Strategies of pursuit-evasion game based on improved potential field and differential game theory for mobile robots}.
In: \bbtitle{Pro. International Conference on Instrumentation, Measurement, Computer, Communication and Control},
\bconflocation{Harbin, China},
pp. \bfpage{1452}--\blpage{1456}
(\byear{2012})
\end{bchapter}
\endbibitem

\bibitem{chen2020dynamic}
\begin{barticle}
\bauthor{\bsnm{Chen}, \binits{X.}},
\bauthor{\bsnm{Zhao}, \binits{M.}},
\bauthor{\bsnm{Yin}, \binits{L.}}:
\batitle{Dynamic path planning of the uav avoiding static and moving obstacles}.
\bjtitle{Journal of Intelligent \& Robotic Systems}
\bvolume{99},
\bfpage{909}--\blpage{931}
(\byear{2020})
\end{barticle}
\endbibitem

\bibitem{ajeil2021novel}
\begin{barticle}
\bauthor{\bsnm{Ajeil}, \binits{F.H.}},
\bauthor{\bsnm{Ibraheem}, \binits{I.K.}},
\bauthor{\bsnm{Humaidi}, \binits{A.J.}},
\bauthor{\bsnm{Khan}, \binits{Z.H.}}:
\batitle{A novel path planning algorithm for mobile robot in dynamic environments using modified bat swarm optimization}.
\bjtitle{The Journal of Engineering}
\bvolume{2021}(\bissue{1}),
\bfpage{37}--\blpage{48}
(\byear{2021})
\end{barticle}
\endbibitem

\bibitem{zaccone2020collision}
\begin{barticle}
\bauthor{\bsnm{Zaccone}, \binits{R.}},
\bauthor{\bsnm{Martelli}, \binits{M.}}:
\batitle{A collision avoidance algorithm for ship guidance applications}.
\bjtitle{Journal of Marine Engineering \& Technology}
\bvolume{19}(\bissue{sup1}),
\bfpage{62}--\blpage{75}
(\byear{2020})
\end{barticle}
\endbibitem

\bibitem{tang2021gwo}
\begin{barticle}
\bauthor{\bsnm{Tang}, \binits{H.}},
\bauthor{\bsnm{Sun}, \binits{W.}},
\bauthor{\bsnm{Lin}, \binits{A.}},
\bauthor{\bsnm{Xue}, \binits{M.}},
\bauthor{\bsnm{Zhang}, \binits{X.}}:
\batitle{A gwo-based multi-robot cooperation method for target searching in unknown environments}.
\bjtitle{Expert Systems with Applications}
\bvolume{186},
\bfpage{115795}
(\byear{2021})
\end{barticle}
\endbibitem

\bibitem{goodwin2015escape}
\begin{barticle}
\bauthor{\bsnm{Goodwin}, \binits{M.}},
\bauthor{\bsnm{Granmo}, \binits{O.-C.}},
\bauthor{\bsnm{Radianti}, \binits{J.}}:
\batitle{Escape planning in realistic fire scenarios with ant colony optimisation}.
\bjtitle{Applied Intelligence}
\bvolume{42},
\bfpage{24}--\blpage{35}
(\byear{2015})
\end{barticle}
\endbibitem

\bibitem{zhu2020receding}
\begin{barticle}
\bauthor{\bsnm{Zhu}, \binits{Q.}},
\bauthor{\bsnm{Wang}, \binits{K.}},
\bauthor{\bsnm{Shao}, \binits{Z.}},
\bauthor{\bsnm{Biegler}, \binits{L.T.}}:
\batitle{Receding horizon optimization method for solving the cops and robbers problems in a complex environment with obstacles: categories (2),(5)}.
\bjtitle{Journal of Intelligent \& Robotic Systems}
\bvolume{100},
\bfpage{83}--\blpage{112}
(\byear{2020})
\end{barticle}
\endbibitem

\bibitem{cognetti2017real}
\begin{bchapter}
\bauthor{\bsnm{Cognetti}, \binits{M.}},
\bauthor{\bsnm{De~Simone}, \binits{D.}},
\bauthor{\bsnm{Patota}, \binits{F.}},
\bauthor{\bsnm{Scianca}, \binits{N.}},
\bauthor{\bsnm{Lanari}, \binits{L.}},
\bauthor{\bsnm{Oriolo}, \binits{G.}}:
\bctitle{Real-time pursuit-evasion with humanoid robots}.
In: \bbtitle{2017 IEEE International Conference on Robotics and Automation (ICRA)},
pp. \bfpage{4090}--\blpage{4095}
(\byear{2017}).
\bcomment{IEEE}
\end{bchapter}
\endbibitem

\bibitem{isaacs1999differential}
\begin{bbook}
\bauthor{\bsnm{Isaacs}, \binits{R.}}:
\bbtitle{Differential Games: a Mathematical Theory with Applications to Warfare and Pursuit, Control and Optimization}.
\bpublisher{Courier Corporation},
\blocation{New York}
(\byear{1999})
\end{bbook}
\endbibitem

\bibitem{liang2023collaborative}
\begin{botherref}
\oauthor{\bsnm{Liang}, \binits{X.}},
\oauthor{\bsnm{Zhou}, \binits{B.}},
\oauthor{\bsnm{Jiang}, \binits{L.}},
\oauthor{\bsnm{Meng}, \binits{G.}},
\oauthor{\bsnm{Xiu}, \binits{Y.}}:
Collaborative pursuit-evasion game of multi-uavs based on apollonius circle in the environment with obstacle.
Connection Science,
1--21
(2023)
\end{botherref}
\endbibitem

\bibitem{liang2019multi}
\begin{bchapter}
\bauthor{\bsnm{Liang}, \binits{L.}},
\bauthor{\bsnm{Deng}, \binits{F.}},
\bauthor{\bsnm{Shi}, \binits{X.}},
\bauthor{\bsnm{Lu}, \binits{M.}}:
\bctitle{A multi-robot cooperative confrontation game with limited range of motion}.
In: \bbtitle{2019 IEEE International Conference on Robotics and Biomimetics (ROBIO)},
pp. \bfpage{764}--\blpage{769}
(\byear{2019}).
\bcomment{IEEE}
\end{bchapter}
\endbibitem

\bibitem{ramana2017pursuit}
\begin{barticle}
\bauthor{\bsnm{Ramana}, \binits{M.V.}},
\bauthor{\bsnm{Kothari}, \binits{M.}}:
\batitle{Pursuit-evasion games of high speed evader}.
\bjtitle{Journal of intelligent \& robotic systems}
\bvolume{85},
\bfpage{293}--\blpage{306}
(\byear{2017})
\end{barticle}
\endbibitem

\bibitem{sun2022cooperative}
\begin{barticle}
\bauthor{\bsnm{Sun}, \binits{Z.}},
\bauthor{\bsnm{Sun}, \binits{H.}},
\bauthor{\bsnm{Li}, \binits{P.}},
\bauthor{\bsnm{Zou}, \binits{J.}}:
\batitle{Cooperative strategy for pursuit-evasion problem with collision avoidance}.
\bjtitle{Ocean Engineering}
\bvolume{266},
\bfpage{112742}
(\byear{2022})
\end{barticle}
\endbibitem

\bibitem{kumar2019experimental}
\begin{barticle}
\bauthor{\bsnm{Kumar}, \binits{A.}},
\bauthor{\bsnm{Ojha}, \binits{A.}}:
\batitle{Experimental evaluation of certain pursuit and evasion schemes for wheeled mobile robots}.
\bjtitle{International Journal of Automation and Computing}
\bvolume{16}(\bissue{4}),
\bfpage{491}--\blpage{510}
(\byear{2019})
\end{barticle}
\endbibitem

\bibitem{ruiz2023time}
\begin{barticle}
\bauthor{\bsnm{Ruiz}, \binits{U.}}:
\batitle{Time-optimal escape of an omnidirectional agent from the field of view of a differential drive robot}.
\bjtitle{International Journal of Control, Automation and Systems}
\bvolume{21}(\bissue{1}),
\bfpage{292}--\blpage{305}
(\byear{2023})
\end{barticle}
\endbibitem

\bibitem{oyler2016pursuit}
\begin{barticle}
\bauthor{\bsnm{Oyler}, \binits{D.W.}},
\bauthor{\bsnm{Kabamba}, \binits{P.T.}},
\bauthor{\bsnm{Girard}, \binits{A.R.}}:
\batitle{Pursuit--evasion games in the presence of obstacles}.
\bjtitle{Automatica}
\bvolume{65},
\bfpage{1}--\blpage{11}
(\byear{2016})
\end{barticle}
\endbibitem

\bibitem{yan2018reach}
\begin{barticle}
\bauthor{\bsnm{Yan}, \binits{R.}},
\bauthor{\bsnm{Shi}, \binits{Z.}},
\bauthor{\bsnm{Zhong}, \binits{Y.}}:
\batitle{Reach-avoid games with two defenders and one attacker: An analytical approach}.
\bjtitle{IEEE transactions on cybernetics}
\bvolume{49}(\bissue{3}),
\bfpage{1035}--\blpage{1046}
(\byear{2018})
\end{barticle}
\endbibitem

\bibitem{fang2020cooperative}
\begin{barticle}
\bauthor{\bsnm{Fang}, \binits{X.}},
\bauthor{\bsnm{Wang}, \binits{C.}},
\bauthor{\bsnm{Xie}, \binits{L.}},
\bauthor{\bsnm{Chen}, \binits{J.}}:
\batitle{Cooperative pursuit with multi-pursuer and one faster free-moving evader}.
\bjtitle{IEEE Transactions on Cybernetics}
\bvolume{52}(\bissue{3}),
\bfpage{1405}--\blpage{1414}
(\byear{2020})
\end{barticle}
\endbibitem

\bibitem{di2019optimizing}
\begin{barticle}
\bauthor{\bsnm{Di}, \binits{K.}},
\bauthor{\bsnm{Yang}, \binits{S.}},
\bauthor{\bsnm{Wang}, \binits{W.}},
\bauthor{\bsnm{Yan}, \binits{F.}},
\bauthor{\bsnm{Xing}, \binits{H.}},
\bauthor{\bsnm{Jiang}, \binits{J.}},
\bauthor{\bsnm{Jiang}, \binits{Y.}}:
\batitle{Optimizing evasive strategies for an evader with imperfect vision capacity}.
\bjtitle{Journal of Intelligent \& Robotic Systems}
\bvolume{96}(\bissue{3}),
\bfpage{419}--\blpage{437}
(\byear{2019})
\end{barticle}
\endbibitem

\bibitem{wang2019fuzzy}
\begin{barticle}
\bauthor{\bsnm{Wang}, \binits{L.}},
\bauthor{\bsnm{Wang}, \binits{M.}},
\bauthor{\bsnm{Yue}, \binits{T.}}:
\batitle{A fuzzy deterministic policy gradient algorithm for pursuit-evasion differential games}.
\bjtitle{Neurocomputing}
\bvolume{362},
\bfpage{106}--\blpage{117}
(\byear{2019})
\end{barticle}
\endbibitem

\bibitem{weitzenfeld2008prey}
\begin{barticle}
\bauthor{\bsnm{Weitzenfeld}, \binits{A.}}:
\batitle{A prey catching and predator avoidance neural-schema architecture for single and multiple robots}.
\bjtitle{Journal of Intelligent and Robotic Systems}
\bvolume{51},
\bfpage{203}--\blpage{233}
(\byear{2008})
\end{barticle}
\endbibitem

\bibitem{qi2020deep}
\begin{bchapter}
\bauthor{\bsnm{Qi}, \binits{Q.}},
\bauthor{\bsnm{Zhang}, \binits{X.}},
\bauthor{\bsnm{Guo}, \binits{X.}}:
\bctitle{A deep reinforcement learning approach for the pursuit evasion game in the presence of obstacles}.
In: \bbtitle{Proc. IEEE International Conference on Real-time Computing and Robotics},
\bconflocation{Asahikawa, Japan},
pp. \bfpage{68}--\blpage{73}
(\byear{2020})
\end{bchapter}
\endbibitem

\bibitem{zhang2022game}
\begin{botherref}
\oauthor{\bsnm{Zhang}, \binits{R.}},
\oauthor{\bsnm{Zong}, \binits{Q.}},
\oauthor{\bsnm{Zhang}, \binits{X.}},
\oauthor{\bsnm{Dou}, \binits{L.}},
\oauthor{\bsnm{Tian}, \binits{B.}}:
Game of drones: multi-uav pursuit-evasion game with online motion planning by deep reinforcement learning.
IEEE Transactions on Neural Networks and Learning Systems
(2022)
\end{botherref}
\endbibitem

\bibitem{xu2022autonomous}
\begin{barticle}
\bauthor{\bsnm{Xu}, \binits{G.}},
\bauthor{\bsnm{Jiang}, \binits{W.}},
\bauthor{\bsnm{Wang}, \binits{Z.}},
\bauthor{\bsnm{Wang}, \binits{Y.}}:
\batitle{Autonomous obstacle avoidance and target tracking of uav based on deep reinforcement learning}.
\bjtitle{Journal of Intelligent \& Robotic Systems}
\bvolume{104}(\bissue{4}),
\bfpage{60}
(\byear{2022})
\end{barticle}
\endbibitem

\bibitem{xu2022pursuit}
\begin{barticle}
\bauthor{\bsnm{Xu}, \binits{C.}},
\bauthor{\bsnm{Zhang}, \binits{Y.}},
\bauthor{\bsnm{Wang}, \binits{W.}},
\bauthor{\bsnm{Dong}, \binits{L.}}:
\batitle{Pursuit and evasion strategy of a differential game based on deep reinforcement learning}.
\bjtitle{Frontiers in Bioengineering and Biotechnology}
\bvolume{10},
\bfpage{827408}
(\byear{2022})
\end{barticle}
\endbibitem

\bibitem{ji2021obstacle}
\begin{barticle}
\bauthor{\bsnm{Ji}, \binits{X.}},
\bauthor{\bsnm{Hai}, \binits{J.}},
\bauthor{\bsnm{Luo}, \binits{W.}},
\bauthor{\bsnm{Lin}, \binits{C.}},
\bauthor{\bsnm{Xiong}, \binits{Y.}},
\bauthor{\bsnm{Ou}, \binits{Z.}},
\bauthor{\bsnm{Wen}, \binits{J.}}:
\batitle{Obstacle avoidance in multi-agent formation process based on deep reinforcement learning}.
\bjtitle{Journal of Shanghai Jiaotong University (Science)}
\bvolume{26},
\bfpage{680}--\blpage{685}
(\byear{2021})
\end{barticle}
\endbibitem

\bibitem{gao2019fast}
\begin{bchapter}
\bauthor{\bsnm{Gao}, \binits{Y.}},
\bauthor{\bsnm{Sibirtseva}, \binits{E.}},
\bauthor{\bsnm{Castellano}, \binits{G.}},
\bauthor{\bsnm{Kragic}, \binits{D.}}:
\bctitle{Fast adaptation with meta-reinforcement learning for trust modelling in human-robot interaction}.
In: \bbtitle{2019 IEEE/RSJ International Conference on Intelligent Robots and Systems (IROS)},
pp. \bfpage{305}--\blpage{312}
(\byear{2019}).
\bcomment{IEEE}
\end{bchapter}
\endbibitem

\bibitem{shree2022learning}
\begin{bchapter}
\bauthor{\bsnm{Shree}, \binits{V.}},
\bauthor{\bsnm{Allen}, \binits{S.}},
\bauthor{\bsnm{Asfora}, \binits{B.}},
\bauthor{\bsnm{Banfi}, \binits{J.}},
\bauthor{\bsnm{Campbell}, \binits{M.}}:
\bctitle{Learning to assess danger from movies for cooperative escape planning in hazardous environments}.
In: \bbtitle{2022 IEEE/RSJ International Conference on Intelligent Robots and Systems (IROS)},
pp. \bfpage{5885}--\blpage{5892}
(\byear{2022}).
\bcomment{IEEE}
\end{bchapter}
\endbibitem

\bibitem{Li2021bio}
\begin{barticle}
\bauthor{\bsnm{Li}, \binits{J.}},
\bauthor{\bsnm{Xu}, \binits{Z.}},
\bauthor{\bsnm{Zhu}, \binits{D.}},
\bauthor{\bsnm{Dong}, \binits{K.}},
\bauthor{\bsnm{Yan}, \binits{T.}},
\bauthor{\bsnm{Zeng}, \binits{Z.}},
\bauthor{\bsnm{Yang}, \binits{S.X.}}:
\batitle{Bio-inspired intelligence with applications to robotics: a survey}.
\bjtitle{Intelligence \& Robotics}
\bvolume{1}(\bissue{1}),
\bfpage{58}--\blpage{83}
(\byear{2021})
\end{barticle}
\endbibitem

\bibitem{tu2021bio}
\begin{barticle}
\bauthor{\bsnm{Tu}, \binits{Z.}},
\bauthor{\bsnm{Fei}, \binits{F.}},
\bauthor{\bsnm{Deng}, \binits{X.}}:
\batitle{Bio-inspired rapid escape and tight body flip on an at-scale flapping wing hummingbird robot via reinforcement learning}.
\bjtitle{IEEE Transactions on Robotics}
\bvolume{37}(\bissue{5}),
\bfpage{1742}--\blpage{1751}
(\byear{2021})
\end{barticle}
\endbibitem

\bibitem{prasath2022dynamics}
\begin{barticle}
\bauthor{\bsnm{Prasath}, \binits{S.G.}},
\bauthor{\bsnm{Mandal}, \binits{S.}},
\bauthor{\bsnm{Giardina}, \binits{F.}},
\bauthor{\bsnm{Kennedy}, \binits{J.}},
\bauthor{\bsnm{Murthy}, \binits{V.N.}},
\bauthor{\bsnm{Mahadevan}, \binits{L.}}:
\batitle{Dynamics of cooperative excavation in ant and robot collectives}.
\bjtitle{Elife}
\bvolume{11},
\bfpage{79638}
(\byear{2022})
\end{barticle}
\endbibitem

\bibitem{lin2011goqbot}
\begin{barticle}
\bauthor{\bsnm{Lin}, \binits{H.-T.}},
\bauthor{\bsnm{Leisk}, \binits{G.G.}},
\bauthor{\bsnm{Trimmer}, \binits{B.}}:
\batitle{Goqbot: a caterpillar-inspired soft-bodied rolling robot}.
\bjtitle{Bioinspiration \& biomimetics}
\bvolume{6}(\bissue{2}),
\bfpage{026007}
(\byear{2011})
\end{barticle}
\endbibitem

\bibitem{nishimura2016learning}
\begin{bchapter}
\bauthor{\bsnm{Nishimura}, \binits{Y.}},
\bauthor{\bsnm{Mikami}, \binits{S.}}:
\bctitle{Learning adaptive escape behavior for wheel-legged robot by inner torque informationprasath2022dynamics}.
In: \bbtitle{2016 Joint 8th International Conference on Soft Computing and Intelligent Systems (SCIS) and 17th International Symposium on Advanced Intelligent Systems (ISIS)},
pp. \bfpage{10}--\blpage{15}
(\byear{2016}).
\bcomment{IEEE}
\end{bchapter}
\endbibitem

\bibitem{weymouth2015ultra}
\begin{barticle}
\bauthor{\bsnm{Weymouth}, \binits{G.}},
\bauthor{\bsnm{Subramaniam}, \binits{V.}},
\bauthor{\bsnm{Triantafyllou}, \binits{M.}}:
\batitle{Ultra-fast escape maneuver of an octopus-inspired robot}.
\bjtitle{Bioinspiration \& biomimetics}
\bvolume{10}(\bissue{1}),
\bfpage{016016}
(\byear{2015})
\end{barticle}
\endbibitem

\bibitem{he2023copebot}
\begin{barticle}
\bauthor{\bsnm{He}, \binits{Z.}},
\bauthor{\bsnm{Yang}, \binits{Y.}},
\bauthor{\bsnm{Jiao}, \binits{P.}},
\bauthor{\bsnm{Wang}, \binits{H.}},
\bauthor{\bsnm{Lin}, \binits{G.}},
\bauthor{\bsnm{P{\"a}htz}, \binits{T.}}:
\batitle{Copebot: underwater soft robot with copepod-like locomotion}.
\bjtitle{Soft Robotics}
\bvolume{10}(\bissue{2}),
\bfpage{314}--\blpage{325}
(\byear{2023})
\end{barticle}
\endbibitem

\bibitem{berlinger2021self}
\begin{bchapter}
\bauthor{\bsnm{Berlinger}, \binits{F.}},
\bauthor{\bsnm{Wulkop}, \binits{P.}},
\bauthor{\bsnm{Nagpal}, \binits{R.}}:
\bctitle{Self-organized evasive fountain maneuvers with a bioinspired underwater robot collective}.
In: \bbtitle{IEEE International Conference on Robotics and Automation},
\bconflocation{Xi'an, China},
pp. \bfpage{9204}--\blpage{9211}
(\byear{2021})
\end{bchapter}
\endbibitem

\bibitem{berlinger2021implicit}
\begin{barticle}
\bauthor{\bsnm{Berlinger}, \binits{F.}},
\bauthor{\bsnm{Gauci}, \binits{M.}},
\bauthor{\bsnm{Nagpal}, \binits{R.}}:
\batitle{Implicit coordination for 3d underwater collective behaviors in a fish-inspired robot swarm}.
\bjtitle{Science Robotics}
\bvolume{6}(\bissue{50}),
\bfpage{8668}
(\byear{2021})
\end{barticle}
\endbibitem

\bibitem{min2011design}
\begin{bchapter}
\bauthor{\bsnm{Min}, \binits{H.}},
\bauthor{\bsnm{Wang}, \binits{Z.}}:
\bctitle{Design and analysis of group escape behavior for distributed autonomous mobile robots}.
In: \bbtitle{IEEE International Conference on Robotics and Automation},
\bconflocation{Shanghai, China},
pp. \bfpage{6128}--\blpage{6135}
(\byear{2011})
\end{bchapter}
\endbibitem

\bibitem{marchese2014autonomous}
\begin{barticle}
\bauthor{\bsnm{Marchese}, \binits{A.D.}},
\bauthor{\bsnm{Onal}, \binits{C.D.}},
\bauthor{\bsnm{Rus}, \binits{D.}}:
\batitle{Autonomous soft robotic fish capable of escape maneuvers using fluidic elastomer actuators}.
\bjtitle{Soft robotics}
\bvolume{1}(\bissue{1}),
\bfpage{75}--\blpage{87}
(\byear{2014})
\end{barticle}
\endbibitem

\bibitem{papadopoulou2022self}
\begin{barticle}
\bauthor{\bsnm{Papadopoulou}, \binits{M.}},
\bauthor{\bsnm{Hildenbrandt}, \binits{H.}},
\bauthor{\bsnm{Sankey}, \binits{D.W.}},
\bauthor{\bsnm{Portugal}, \binits{S.J.}},
\bauthor{\bsnm{Hemelrijk}, \binits{C.K.}}:
\batitle{Self-organization of collective escape in pigeon flocks}.
\bjtitle{PLoS Computational Biology}
\bvolume{18}(\bissue{1}),
\bfpage{1009772}
(\byear{2022})
\end{barticle}
\endbibitem

\bibitem{doran2022fish}
\begin{barticle}
\bauthor{\bsnm{Doran}, \binits{C.}},
\bauthor{\bsnm{Bierbach}, \binits{D.}},
\bauthor{\bsnm{Lukas}, \binits{J.}},
\bauthor{\bsnm{Klamser}, \binits{P.}},
\bauthor{\bsnm{Landgraf}, \binits{T.}},
\bauthor{\bsnm{Klenz}, \binits{H.}},
\bauthor{\bsnm{Habedank}, \binits{M.}},
\bauthor{\bsnm{Arias-Rodriguez}, \binits{L.}},
\bauthor{\bsnm{Krause}, \binits{S.}},
\bauthor{\bsnm{Romanczuk}, \binits{P.}}, \betal:
\batitle{Fish waves as emergent collective antipredator behavior}.
\bjtitle{Current Biology}
\bvolume{32}(\bissue{3}),
\bfpage{708}--\blpage{714}
(\byear{2022})
\end{barticle}
\endbibitem

\bibitem{ioannou2011social}
\begin{barticle}
\bauthor{\bsnm{Ioannou}, \binits{C.C.}},
\bauthor{\bsnm{Couzin}, \binits{I.D.}},
\bauthor{\bsnm{James}, \binits{R.}},
\bauthor{\bsnm{Croft}, \binits{D.P.}},
\bauthor{\bsnm{Krause}, \binits{J.}}:
\batitle{Social organisation and information transfer in schooling fish}.
\bjtitle{Fish cognition and behavior}
\bvolume{2},
\bfpage{217}--\blpage{239}
(\byear{2011})
\end{barticle}
\endbibitem

\bibitem{ishiwaka2021foids}
\begin{barticle}
\bauthor{\bsnm{Ishiwaka}, \binits{Y.}},
\bauthor{\bsnm{Zeng}, \binits{X.S.}},
\bauthor{\bsnm{Eastman}, \binits{M.L.}},
\bauthor{\bsnm{Kakazu}, \binits{S.}},
\bauthor{\bsnm{Gross}, \binits{S.}},
\bauthor{\bsnm{Mizutani}, \binits{R.}},
\bauthor{\bsnm{Nakada}, \binits{M.}}:
\batitle{Foids: bio-inspired fish simulation for generating synthetic datasets}.
\bjtitle{ACM Transactions on Graphics (TOG)}
\bvolume{40}(\bissue{6}),
\bfpage{1}--\blpage{15}
(\byear{2021})
\end{barticle}
\endbibitem

\bibitem{han2022snake}
\begin{barticle}
\bauthor{\bsnm{Han}, \binits{S.}},
\bauthor{\bsnm{Chon}, \binits{S.}},
\bauthor{\bsnm{Kim}, \binits{J.}},
\bauthor{\bsnm{Seo}, \binits{J.}},
\bauthor{\bsnm{Shin}, \binits{D.G.}},
\bauthor{\bsnm{Park}, \binits{S.}},
\bauthor{\bsnm{Kim}, \binits{J.T.}},
\bauthor{\bsnm{Kim}, \binits{J.}},
\bauthor{\bsnm{Jin}, \binits{M.}},
\bauthor{\bsnm{Cho}, \binits{J.}}:
\batitle{Snake robot gripper module for search and rescue in narrow spaces}.
\bjtitle{IEEE Robotics and Automation Letters}
\bvolume{7}(\bissue{2}),
\bfpage{1667}--\blpage{1673}
(\byear{2022})
\end{barticle}
\endbibitem

\bibitem{sun2021bit}
\begin{barticle}
\bauthor{\bsnm{Sun}, \binits{Z.}},
\bauthor{\bsnm{Yang}, \binits{H.}},
\bauthor{\bsnm{Ma}, \binits{Y.}},
\bauthor{\bsnm{Wang}, \binits{X.}},
\bauthor{\bsnm{Mo}, \binits{Y.}},
\bauthor{\bsnm{Li}, \binits{H.}},
\bauthor{\bsnm{Jiang}, \binits{Z.}}:
\batitle{Bit-dmr: A humanoid dual-arm mobile robot for complex rescue operations}.
\bjtitle{IEEE Robotics and Automation Letters}
\bvolume{7}(\bissue{2}),
\bfpage{802}--\blpage{809}
(\byear{2021})
\end{barticle}
\endbibitem

\bibitem{boukas2014robot}
\begin{barticle}
\bauthor{\bsnm{Boukas}, \binits{E.}},
\bauthor{\bsnm{Kostavelis}, \binits{I.}},
\bauthor{\bsnm{Gasteratos}, \binits{A.}},
\bauthor{\bsnm{Sirakoulis}, \binits{G.C.}}:
\batitle{Robot guided crowd evacuation}.
\bjtitle{IEEE Transactions on Automation Science and Engineering}
\bvolume{12}(\bissue{2}),
\bfpage{739}--\blpage{751}
(\byear{2014})
\end{barticle}
\endbibitem

\bibitem{lopez2018new}
\begin{barticle}
\bauthor{\bsnm{Lopez}, \binits{T.}}:
\batitle{New smet will take the load off infantry soldiers}.
\bjtitle{Retrieved January}
\bvolume{23},
\bfpage{2019}
(\byear{2018})
\end{barticle}
\endbibitem

\bibitem{kumar2021drone}
\begin{barticle}
\bauthor{\bsnm{Kumar}, \binits{A.}},
\bauthor{\bsnm{Sharma}, \binits{K.}},
\bauthor{\bsnm{Singh}, \binits{H.}},
\bauthor{\bsnm{Naugriya}, \binits{S.G.}},
\bauthor{\bsnm{Gill}, \binits{S.S.}},
\bauthor{\bsnm{Buyya}, \binits{R.}}:
\batitle{A drone-based networked system and methods for combating coronavirus disease (covid-19) pandemic}.
\bjtitle{Future Generation Computer Systems}
\bvolume{115},
\bfpage{1}--\blpage{19}
(\byear{2021})
\end{barticle}
\endbibitem

\bibitem{drew2021multi}
\begin{barticle}
\bauthor{\bsnm{Drew}, \binits{D.S.}}:
\batitle{Multi-agent systems for search and rescue applications}.
\bjtitle{Current Robotics Reports}
\bvolume{2},
\bfpage{189}--\blpage{200}
(\byear{2021})
\end{barticle}
\endbibitem

\bibitem{wang2020three}
\begin{barticle}
\bauthor{\bsnm{Wang}, \binits{H.}},
\bauthor{\bsnm{Zhang}, \binits{C.}},
\bauthor{\bsnm{Song}, \binits{Y.}},
\bauthor{\bsnm{Pang}, \binits{B.}},
\bauthor{\bsnm{Zhang}, \binits{G.}}:
\batitle{Three-dimensional reconstruction based on visual slam of mobile robot in search and rescue disaster scenarios}.
\bjtitle{Robotica}
\bvolume{38}(\bissue{2}),
\bfpage{350}--\blpage{373}
(\byear{2020})
\end{barticle}
\endbibitem

\bibitem{dong2021uav}
\begin{barticle}
\bauthor{\bsnm{Dong}, \binits{J.}},
\bauthor{\bsnm{Ota}, \binits{K.}},
\bauthor{\bsnm{Dong}, \binits{M.}}:
\batitle{Uav-based real-time survivor detection system in post-disaster search and rescue operations}.
\bjtitle{IEEE Journal on Miniaturization for Air and Space Systems}
\bvolume{2}(\bissue{4}),
\bfpage{209}--\blpage{219}
(\byear{2021})
\end{barticle}
\endbibitem

\bibitem{zhou2022robot}
\begin{barticle}
\bauthor{\bsnm{Zhou}, \binits{M.}},
\bauthor{\bsnm{Dong}, \binits{H.}},
\bauthor{\bsnm{Ge}, \binits{S.}},
\bauthor{\bsnm{Wang}, \binits{X.}},
\bauthor{\bsnm{Wang}, \binits{F.-Y.}}:
\batitle{Robot-guided crowd evacuation in a railway hub station in case of emergencies}.
\bjtitle{Journal of Intelligent \& Robotic Systems}
\bvolume{104}(\bissue{4}),
\bfpage{67}
(\byear{2022})
\end{barticle}
\endbibitem

\bibitem{zhou2019guided}
\begin{barticle}
\bauthor{\bsnm{Zhou}, \binits{M.}},
\bauthor{\bsnm{Dong}, \binits{H.}},
\bauthor{\bsnm{Ioannou}, \binits{P.A.}},
\bauthor{\bsnm{Zhao}, \binits{Y.}},
\bauthor{\bsnm{Wang}, \binits{F.-Y.}}:
\batitle{Guided crowd evacuation: approaches and challenges}.
\bjtitle{IEEE/CAA Journal of Automatica Sinica}
\bvolume{6}(\bissue{5}),
\bfpage{1081}--\blpage{1094}
(\byear{2019})
\end{barticle}
\endbibitem

\bibitem{sakour2017robot}
\begin{barticle}
\bauthor{\bsnm{Sakour}, \binits{I.}},
\bauthor{\bsnm{Hu}, \binits{H.}}:
\batitle{Robot-assisted crowd evacuation under emergency situations: A survey}.
\bjtitle{Robotics}
\bvolume{6}(\bissue{2}),
\bfpage{8}
(\byear{2017})
\end{barticle}
\endbibitem

\bibitem{tang2016human}
\begin{barticle}
\bauthor{\bsnm{Tang}, \binits{B.}},
\bauthor{\bsnm{Jiang}, \binits{C.}},
\bauthor{\bsnm{He}, \binits{H.}},
\bauthor{\bsnm{Guo}, \binits{Y.}}:
\batitle{Human mobility modeling for robot-assisted evacuation in complex indoor environments}.
\bjtitle{IEEE Transactions on Human-Machine Systems}
\bvolume{46}(\bissue{5}),
\bfpage{694}--\blpage{707}
(\byear{2016})
\end{barticle}
\endbibitem

\bibitem{jiang2017learning}
\begin{barticle}
\bauthor{\bsnm{Jiang}, \binits{C.}},
\bauthor{\bsnm{Ni}, \binits{Z.}},
\bauthor{\bsnm{Guo}, \binits{Y.}},
\bauthor{\bsnm{He}, \binits{H.}}:
\batitle{Learning human--robot interaction for robot-assisted pedestrian flow optimization}.
\bjtitle{IEEE Transactions on Systems, Man, and Cybernetics: Systems}
\bvolume{49}(\bissue{4}),
\bfpage{797}--\blpage{813}
(\byear{2017})
\end{barticle}
\endbibitem

\bibitem{panesar2019artificial}
\begin{barticle}
\bauthor{\bsnm{Panesar}, \binits{S.}},
\bauthor{\bsnm{Cagle}, \binits{Y.}},
\bauthor{\bsnm{Chander}, \binits{D.}},
\bauthor{\bsnm{Morey}, \binits{J.}},
\bauthor{\bsnm{Fernandez-Miranda}, \binits{J.}},
\bauthor{\bsnm{Kliot}, \binits{M.}}:
\batitle{Artificial intelligence and the future of surgical robotics}.
\bjtitle{Annals of surgery}
\bvolume{270}(\bissue{2}),
\bfpage{223}--\blpage{226}
(\byear{2019})
\end{barticle}
\endbibitem

\bibitem{wang2020use}
\begin{bchapter}
\bauthor{\bsnm{Wang}, \binits{H.}},
\bauthor{\bsnm{Cheng}, \binits{H.}},
\bauthor{\bsnm{Hao}, \binits{H.}}:
\bctitle{The use of unmanned aerial vehicle in military operations}.
In: \bbtitle{Man-Machine-Environment System Engineering: Proceedings of the 20th International Conference on MMESE},
pp. \bfpage{939}--\blpage{945}
(\byear{2020}).
\bcomment{Springer}
\end{bchapter}
\endbibitem

\bibitem{javaid2020robotics}
\begin{barticle}
\bauthor{\bsnm{Javaid}, \binits{M.}},
\bauthor{\bsnm{Haleem}, \binits{A.}},
\bauthor{\bsnm{Vaish}, \binits{A.}},
\bauthor{\bsnm{Vaishya}, \binits{R.}},
\bauthor{\bsnm{Iyengar}, \binits{K.P.}}:
\batitle{Robotics applications in covid-19: A review}.
\bjtitle{Journal of Industrial Integration and Management}
\bvolume{5}(\bissue{04}),
\bfpage{441}--\blpage{451}
(\byear{2020})
\end{barticle}
\endbibitem

\bibitem{javaid2020industry}
\begin{barticle}
\bauthor{\bsnm{Javaid}, \binits{M.}},
\bauthor{\bsnm{Haleem}, \binits{A.}},
\bauthor{\bsnm{Vaishya}, \binits{R.}},
\bauthor{\bsnm{Bahl}, \binits{S.}},
\bauthor{\bsnm{Suman}, \binits{R.}},
\bauthor{\bsnm{Vaish}, \binits{A.}}:
\batitle{Industry 4.0 technologies and their applications in fighting covid-19 pandemic}.
\bjtitle{Diabetes \& Metabolic Syndrome: Clinical Research \& Reviews}
\bvolume{14}(\bissue{4}),
\bfpage{419}--\blpage{422}
(\byear{2020})
\end{barticle}
\endbibitem

\bibitem{kyprianou2022towards}
\begin{barticle}
\bauthor{\bsnm{Kyprianou}, \binits{G.}},
\bauthor{\bsnm{Doitsidis}, \binits{L.}},
\bauthor{\bsnm{Chatzichristofis}, \binits{S.A.}}:
\batitle{Towards the achievement of path planning with multi-robot systems in dynamic environments}.
\bjtitle{Journal of Intelligent \& Robotic Systems}
\bvolume{104}(\bissue{1}),
\bfpage{15}
(\byear{2022})
\end{barticle}
\endbibitem

\bibitem{kobayashi2022local}
\begin{barticle}
\bauthor{\bsnm{Kobayashi}, \binits{M.}},
\bauthor{\bsnm{Motoi}, \binits{N.}}:
\batitle{Local path planning: Dynamic window approach with virtual manipulators considering dynamic obstacles}.
\bjtitle{IEEE Access}
\bvolume{10},
\bfpage{17018}--\blpage{17029}
(\byear{2022})
\end{barticle}
\endbibitem

\bibitem{huang2022sine}
\begin{barticle}
\bauthor{\bsnm{Huang}, \binits{T.}},
\bauthor{\bsnm{Pan}, \binits{H.}},
\bauthor{\bsnm{Sun}, \binits{W.}},
\bauthor{\bsnm{Gao}, \binits{H.}}:
\batitle{Sine resistance network-based motion planning approach for autonomous electric vehicles in dynamic environments}.
\bjtitle{IEEE Transactions on Transportation Electrification}
\bvolume{8}(\bissue{2}),
\bfpage{2862}--\blpage{2873}
(\byear{2022})
\end{barticle}
\endbibitem

\bibitem{fareh2021active}
\begin{barticle}
\bauthor{\bsnm{Fareh}, \binits{R.}},
\bauthor{\bsnm{Khadraoui}, \binits{S.}},
\bauthor{\bsnm{Abdallah}, \binits{M.Y.}},
\bauthor{\bsnm{Baziyad}, \binits{M.}},
\bauthor{\bsnm{Bettayeb}, \binits{M.}}:
\batitle{Active disturbance rejection control for robotic systems: A review}.
\bjtitle{Mechatronics}
\bvolume{80},
\bfpage{102671}
(\byear{2021})
\end{barticle}
\endbibitem

\bibitem{liang2021low}
\begin{barticle}
\bauthor{\bsnm{Liang}, \binits{J.}},
\bauthor{\bsnm{Chen}, \binits{Y.}},
\bauthor{\bsnm{Lai}, \binits{N.}},
\bauthor{\bsnm{He}, \binits{B.}},
\bauthor{\bsnm{Miao}, \binits{Z.}},
\bauthor{\bsnm{Wang}, \binits{Y.}}:
\batitle{Low-complexity prescribed performance control for unmanned aerial manipulator robot system under model uncertainty and unknown disturbances}.
\bjtitle{IEEE Transactions on Industrial Informatics}
\bvolume{18}(\bissue{7}),
\bfpage{4632}--\blpage{4641}
(\byear{2021})
\end{barticle}
\endbibitem

\bibitem{chu2022path}
\begin{botherref}
\oauthor{\bsnm{Chu}, \binits{Z.}},
\oauthor{\bsnm{Wang}, \binits{F.}},
\oauthor{\bsnm{Lei}, \binits{T.}},
\oauthor{\bsnm{Luo}, \binits{C.}}:
Path planning based on deep reinforcement learning for autonomous underwater vehicles under ocean current disturbance.
IEEE Transactions on Intelligent Vehicles
(2022)
\end{botherref}
\endbibitem

\bibitem{ferrer2021secure}
\begin{barticle}
\bauthor{\bsnm{Ferrer}, \binits{E.C.}},
\bauthor{\bsnm{Hardjono}, \binits{T.}},
\bauthor{\bsnm{Pentland}, \binits{A.}},
\bauthor{\bsnm{Dorigo}, \binits{M.}}:
\batitle{Secure and secret cooperation in robot swarms}.
\bjtitle{Science Robotics}
\bvolume{6}(\bissue{56}),
\bfpage{1538}
(\byear{2021})
\end{barticle}
\endbibitem

\bibitem{yang2021review}
\begin{barticle}
\bauthor{\bsnm{Yang}, \binits{C.}},
\bauthor{\bsnm{Zhu}, \binits{Y.}},
\bauthor{\bsnm{Chen}, \binits{Y.}}:
\batitle{A review of human--machine cooperation in the robotics domain}.
\bjtitle{IEEE Transactions on Human-Machine Systems}
\bvolume{52}(\bissue{1}),
\bfpage{12}--\blpage{25}
(\byear{2021})
\end{barticle}
\endbibitem

\bibitem{simetti2021wimust}
\begin{barticle}
\bauthor{\bsnm{Simetti}, \binits{E.}},
\bauthor{\bsnm{Indiveri}, \binits{G.}},
\bauthor{\bsnm{Pascoal}, \binits{A.M.}}:
\batitle{Wimust: A cooperative marine robotic system for autonomous geotechnical surveys}.
\bjtitle{Journal of Field Robotics}
\bvolume{38}(\bissue{2}),
\bfpage{268}--\blpage{288}
(\byear{2021})
\end{barticle}
\endbibitem

\bibitem{zhou2021survey}
\begin{barticle}
\bauthor{\bsnm{Zhou}, \binits{Z.}},
\bauthor{\bsnm{Liu}, \binits{J.}},
\bauthor{\bsnm{Yu}, \binits{J.}}:
\batitle{A survey of underwater multi-robot systems}.
\bjtitle{IEEE/CAA Journal of Automatica Sinica}
\bvolume{9}(\bissue{1}),
\bfpage{1}--\blpage{18}
(\byear{2021})
\end{barticle}
\endbibitem

\bibitem{huang2021novel}
\begin{barticle}
\bauthor{\bsnm{Huang}, \binits{J.}},
\bauthor{\bsnm{Li}, \binits{X.}},
\bauthor{\bsnm{Yang}, \binits{Z.}},
\bauthor{\bsnm{Kong}, \binits{W.}},
\bauthor{\bsnm{Zhao}, \binits{Y.}},
\bauthor{\bsnm{Zhou}, \binits{D.}}:
\batitle{A novel elitism co-evolutionary algorithm for antagonistic weapon-target assignment}.
\bjtitle{IEEE Access}
\bvolume{9},
\bfpage{139668}--\blpage{139684}
(\byear{2021})
\end{barticle}
\endbibitem

\bibitem{agarwal2021implementing}
\begin{barticle}
\bauthor{\bsnm{Agarwal}, \binits{D.}},
\bauthor{\bsnm{Bharti}, \binits{P.S.}}:
\batitle{Implementing modified swarm intelligence algorithm based on slime moulds for path planning and obstacle avoidance problem in mobile robots}.
\bjtitle{Applied Soft Computing}
\bvolume{107},
\bfpage{107372}
(\byear{2021})
\end{barticle}
\endbibitem

\bibitem{moorthy2022distributed}
\begin{barticle}
\bauthor{\bsnm{Moorthy}, \binits{S.}},
\bauthor{\bsnm{Joo}, \binits{Y.H.}}:
\batitle{Distributed leader-following formation control for multiple nonholonomic mobile robots via bioinspired neurodynamic approach}.
\bjtitle{Neurocomputing}
\bvolume{492},
\bfpage{308}--\blpage{321}
(\byear{2022})
\end{barticle}
\endbibitem

\end{thebibliography}

\section*{Declarations}

\subsection{Funding}
This work was supported by Natural Sciences and Engineering Research Council (NSERC) of Canada.

\subsection{Conflict of interest}
The authors have no relevant financial or non-financial interests to disclose.
\subsection{Ethics approval}
Not applicable.
\subsection{Consent to participate}
Not applicable.
\subsection{Consent for publication}
All images cited from others' published works in this paper have been properly acknowledged with corresponding references according to academic standards.
\subsection{Availability of data and materials}
Not applicable.
\subsection{Code availability}
Not applicable.
\subsection{Authors' contributions}
All authors contributed to the idea for the article. The literature search and data analysis were performed by Junfei Li and the critically revised works were performed by Simon X. Yang. All authors read and approved the final manuscript.

\begin{biography}{\includegraphics[width=2cm]{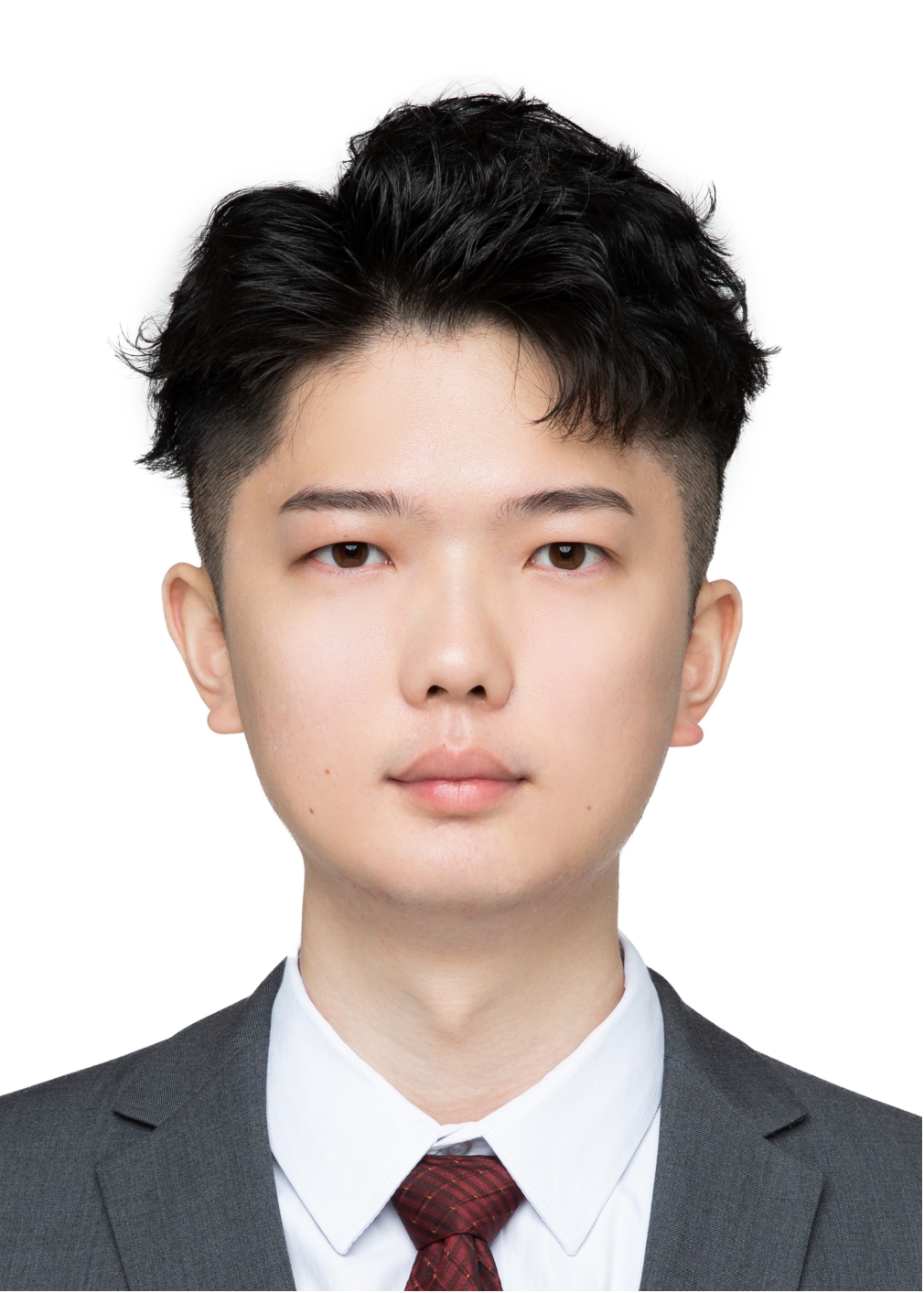}}{%
    \author{Junfei Li} received the B.Eng. degree in communication engineering from Chongqing University of Posts and Telecommunications, Chongqing, China, in 2017, and the Ph.D. degree in engineering systems and computing from the University of Guelph, Ontario, Canada in 2023. He is currently a Postdoctoral Research Fellow at the School of Engineering, University of Guelph, Ontario, Canada. His research interests include escape behaviors in robotics, search and rescue, and bio-inspired algorithms.}
\end{biography}

\begin{biography}{\includegraphics[width=2cm]{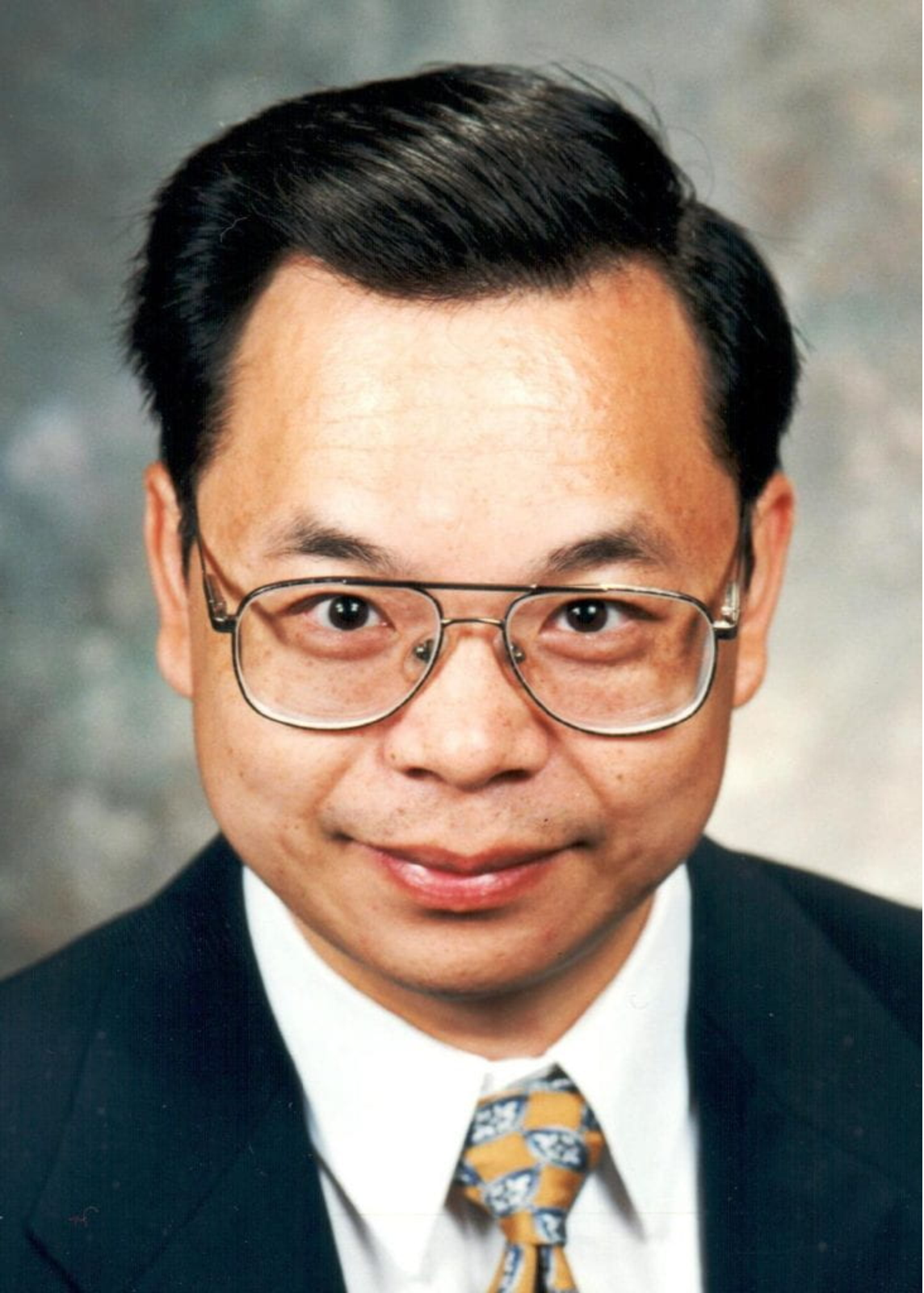}}{%
    \author{Simon X. Yang} (Senior Member, IEEE) received the B.Sc. degree in engineering physics from Beijing University, China in 1987, the first of two M.Sc.  degrees in biophysics from Chinese Academy of Sciences, Beijing, China in 1990, the second M.Sc. degree in electrical engineering from the University of Houston, USA in 1996, and the Ph.D. degree in electrical and computer engineering from the University of Alberta, Edmonton, Canada in 1999. Prof. Yang joined the School of Engineering at the University of Guelph, Canada in 1999. Currently he is a Professor and the Head of the Advanced Robotics \& Intelligent Systems (ARIS) Laboratory at the University of Guelph in Canada. 

Prof. Yang has diversified research expertise. His research interests include intelligent systems, robotics, control systems, sensors and multi-sensor fusion, wireless sensor networks, intelligent communications, intelligent transportation, machine learning, and computational neuroscience. He has published over 550 academic papers, including over 350 journal papers. Prof. Yang he has been very active in professional activities. He serves as the Editor-in-Chief of Intelligence \& Robotics, and International Journal of Robotics and Automation; and an Associate Editor of IEEE Transactions on Cybernetics, IEEE Transactions on Artificial Intelligence, and several other journals. He has been involved in the organization of many international conferences.}
\end{biography}

\end{document}